%% file: main.tex
\documentclass[lettersize,journal]{IEEEtran}
\usepackage{amsmath,amsfonts}
\usepackage{algorithmic}
\usepackage{algorithm}
\usepackage{array}
\usepackage[caption=false,font=normalsize,labelfont=sf,textfont=sf]{subfig}
\usepackage{textcomp}
\usepackage{stfloats}
\usepackage{url}
\usepackage{verbatim}
\usepackage{graphicx}
\usepackage{cite}
\usepackage{multirow}
\usepackage{xspace}
\usepackage{bbm}
\usepackage{booktabs}
\usepackage{color}
\usepackage{setspace}
\usepackage{braket}
\usepackage{tablefootnote}
\usepackage{subcaption}
\usepackage[normalem]{ulem}
\usepackage{hyperref}
\usepackage{cleveref}
\usepackage{makecell}
\usepackage{wrapfig}
\usepackage{sansmath}
\usepackage{float}
\usepackage[table]{xcolor}
\usepackage[most]{tcolorbox}
\usepackage{listings}
\usepackage{tabularx}
\usepackage{pifont}
\newcommand{\ourmethod}{\texttt{UniGFM-RAG}\xspace}

\newcommand{\soc}{\textcolor{black}}
\begin{document}

\title{\ourmethod: Unify Graph Foundation Models and Large Language Models for End-to-End Retrieval and Reasoning}

\author{Zicheng Zhao$^*$, Linhao Luo, Zongji Yu, Junnan Dong, Yanchao Tan, \\ Shirui Pan$^\dagger$, \textit{Senior Member, IEEE}, and Chen Gong$^\dagger$, \textit{Senior Member, IEEE}

% \thanks{Manuscript received April 19, 2021; revised August 16, 2021.}
\thanks{$^*$This work was done while visiting Shanghai Jiao Tong University.}
\thanks{Z. Zhao, Z. Yu and C. Gong are with the School of Automation and Intelligent Sensing, Shanghai Jiao Tong University, Shanghai 200240, China. Email: \{zicheng.zhao99@gmail.com, yuzongji@pjlab.org.cn, chen.gong@sjtu.edu.cn.\}}
\thanks{L. Luo is with the Department of Data Science and AI, Monash University, Melbourne, Australia. E-mail: linhao.luo1@monash.edu.}
\thanks{J. Dong is with the Tencent Youtu Lab, Shanghai 200240, China. E-mail: hansonjdong@tencent.com.}
\thanks{Y. Tan is with School of Computer and Data Science, Fuzhou University, Fuzhou 350116, China. Email: yctan@fzu.edu.cn.}
\thanks{S. Pan is with the School of Information and Communication Technology, Griffith University, Queensland, Australia. Email: s.pan@griffith.edu.au.}
\thanks{$^\dagger$Corresponding Author.}
}

% \author{IEEE Publication Technology,~\IEEEmembership{Staff,~IEEE,}
%         % <-this % stops a space
% \thanks{This paper was produced by the IEEE Publication Technology Group. They are in Piscataway, NJ.}% <-this % stops a space
% \thanks{Manuscript received April 19, 2021; revised August 16, 2021.}}

% The paper headers
\markboth{Journal of \LaTeX\ Class Files,~Vol.~14, No.~8, August~2021}%
{Shell \MakeLowercase{\textit{et al.}}: A Sample Article Using IEEEtran.cls for IEEE Journals}

% \IEEEpubid{0000--0000/00\$00.00~\copyright~2021 IEEE}
% Remember, if you use this you must call \IEEEpubidadjcol in the second
% column for its text to clear the IEEEpubid mark.

\maketitle
\vspace{-.6cm}
\input{sections/00_abstract}
\input{sections/01_introduction}
\input{sections/02_related_work}
\input{sections/04_approach}
\input{sections/05_experiments}
\input{sections/06_conclusions}

% \clearpage

\bibliographystyle{IEEEtran}
\bibliography{main}
\input{sections/biography}
\clearpage
\input{sections/07_appendix}

\end{document}

%% file: sections/00_abstract.tex
\begin{abstract}
Large Language Models (LLMs) have shown remarkable capabilities across tasks, but they still struggle with complex reasoning that requires integrating scattered knowledge. Graph Retrieval-Augmented Generation (GraphRAG) addresses this challenge by organizing fragment knowledge as graphs, while recent Graph Foundation Models (GFMs) further improve GraphRAG with graph-native retrieval and reasoning capabilities. However, existing GraphRAG methods usually adopt decoupled retrieve-then-generate pipelines, which create objective misalignment between retrieval and reasoning, and training GFM retrievers typically depends on expensive human annotations. To address these limitations, we propose \ourmethod, a unified end-to-end GraphRAG framework that tightly couples a GFM retriever with an LLM generator. Given a query, the GFM retriever first reasons over the graph to score and retrieve relevant document nodes. These node embeddings are then passed through a fully differentiable graph bridge, which projects them into soft-weighted graph prefixes according to retrieval scores and combines them with raw document texts to form graph-augmented prompts. In this way, structural information is injected directly into LLM generation, while the final answer loss can be back-propagated through the bridge to jointly optimize retrieval and reasoning. To stabilize training and reduce reliance on costly annotations, we further introduce a semantic-guided distillation module that provides auxiliary supervision using pseudo-labels derived from query-document semantic similarities. Extensive experiments on three multi-hop QA datasets and three out-of-domain GraphRAG benchmarks show that \ourmethod consistently outperforms strong baseline methods and generalizes well, which demonstrates the effectiveness of end-to-end GraphRAG with tightly integrated GFM retrieval and LLM reasoning.
\end{abstract}

\begin{IEEEkeywords}
Large Language Models, Graph Retrieval-Augmented Generation, Graph Foundation Models
\end{IEEEkeywords}
\vspace{-.3cm}

%% file: sections/01_introduction.tex
\section{Introduction}
\begin{figure}[t]
    \centering
    \includegraphics[width=1\columnwidth]{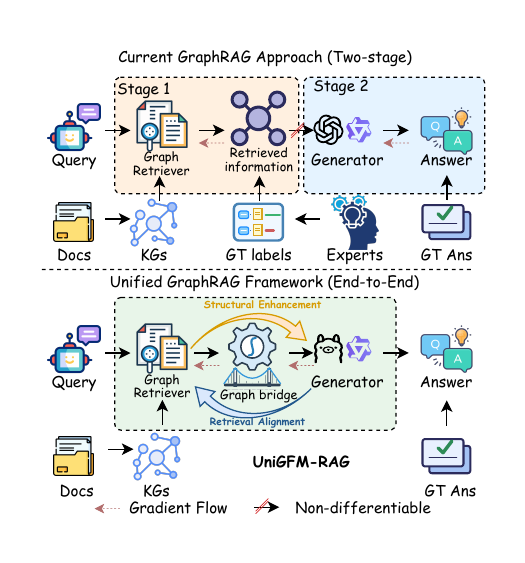}
    \vspace{-.6cm}
    \caption{Comparison between the current two-stage GraphRAG and the proposed \ourmethod. Existing GraphRAG methods usually employ a decoupled retrieve-then-generate pipeline, which often results in objective misalignment. In contrast, \ourmethod introduces a unified end-to-end GraphRAG framework that tightly couples a Graph Foundation Model (GFM) retriever with a Large Language Model (LLM) generator.}
    \label{fig:intro}
    \vspace{-.6cm}
\end{figure}

Large Language Models (LLMs) have demonstrated remarkable performance across a wide range of natural language processing tasks~\cite{achiam2023gpt,yang2025qwen3_tech,liu2024deepseek,dingcodekdd,dullmdata}. However, they still struggle with knowledge-intensive reasoning, where the evidence is often distributed across multiple documents~\cite{listructrag}. Graphs~\cite{pan2024unifying,sun2023all,yanggraphood,jianggram} offer a natural and flexible way of modeling structural information, which makes them particularly suitable for capturing the dependencies within knowledge for complex reasoning. 
Motivated by this, Graph Retrieval-Augmented Generation (GraphRAG)~\cite{peng2025graph,zhang2025survey,huangketrag,zhao2026hyperreco} has emerged as a promising paradigm for enhancing LLM reasoning with graph-structured knowledge. As illustrated in the upper part of Figure~\ref{fig:intro}, a typical GraphRAG system first constructs a graph from the external corpus, such as a knowledge graph~\cite{edge2024local} or a document graph~\cite{wang2024knowledge}, where nodes correspond to documents or entities, and edges encode their relationships. It then employs a graph retriever to navigate the graph and identify relevant nodes, whose associated information is provided to an LLM generator to produce the final answer.

Early graph retrievers relied on the internal capabilities of LLMs to traverse graphs using heuristic search~\cite{jimenez2024hipporag}, global planning~\cite{li2025framework}, or multi-step agentic reasoning~\cite{mathink,weimardkt}. However, these approaches often suffer from inefficiency and limited structural understanding due to the unstructured nature of LLMs. Recently, Graph Foundation Models (GFMs)~\cite{mao2024position,liu2025graph,yanggfmkdd,fang2026unraveling} have demonstrated strong graph reasoning and generalized abilities, which inspires a new wave of applying them to GraphRAG~\cite{luo2025gfm}. These methods utilize graph neural network (GNN)-based GFMs as native graph retrievers to capture intricate dependencies among nodes, which significantly enhances retrieval performance. Despite the promising results, \emph{training GFMs requires large-scale, high-quality relevance labels, which are labor-intensive and difficult to obtain at scale} (\textbf{Limitation 1}).

Beyond the challenge of data acquisition, the architecture of existing GraphRAG frameworks poses another major bottleneck. Most GraphRAG methods adopt a decoupled, two-stage ``retrieve-then-generate'' pipeline~\cite{yu2024rankrag,dong2024don,luo2025gfm}. In this paradigm, the retriever and generator are optimized separately for different objectives: the retriever is trained to maximize retrieval performance rather than improve final answer accuracy. This leads to significant objective misalignment, where \emph{information considered highly relevant by the retriever may be suboptimal or redundant for the reasoning process of the generator} (\textbf{Limitation 2}). Furthermore, because document retrieval is a discrete operation, it prevents continuous gradient backpropagation from the generator to the retriever, which makes joint training even more difficult. Therefore, it is crucial to tightly integrate the GFM retriever and LLM generator into a unified framework that allows end-to-end optimization.

% Instead of facilitating joint optimization, the integration of GFM and LLM into a unified framework also enhance the LLM's ability to leverage graph structure during reasoning. Recent works have explored to adopt GNN to encode the graph structure and inject GNN representations into the LLM as graph-aware prompts~\cite{tang2024graphgpt,chai2025graphllm}, which establish a connection between graph structure and text generation. However, these methods are designed for standalone graph reasoning tasks (e.g., node classification, link prediction \cite{chenllaga}) and thus \emph{do not consider the retriever-generator synergy as well as the cross knowledge dependencies in the context of GraphRAG} (\textbf{Limitation 3}). 

Beyond enabling joint optimization, a unified GFM-LLM framework can further improve the ability of LLM to leverage graph structure during reasoning~\cite{jin2024large}. Prior work has attempted to inject graph structural signals into LLMs through GNN-based graph encoders and graph-aware prompting~\cite{tang2024graphgpt,chai2025graphllm,chen2024graphwiz}. Although effective for standalone graph tasks such as anomaly detection~\cite{qiao2025anomalygfm}, node classification and link prediction~\cite{chenllaga}, these methods assume a pre-defined graph as input and are not designed for retrieval-augmented settings. As a result, they \emph{neither address how to retrieve query-relevant information from external graph nor capture the interaction between graph retrieval and answer generation} (\textbf{Limitation 3}), which is essential in GraphRAG.

% To overcome this architectural misalignment, several methods in conventional RAG~\cite{guu2020retrieval,lewis2020retrieval,singh2021end} have explored end-to-end joint training, where the generator provides feedback on retrieved passages and guides retriever learning. However, these approaches treat retrieved items as independent, scattered evidences and thus largely ignore the complex, multi-document structural relationships that graph encode. Moreover, the retrievers and generators are still designed separately, which leads to a \emph{lack of synergy between the structural understanding of GFM retriever and the reasoning capabilities of LLM generator} (\textbf{Limitation 3}).

In response to these challenges, we propose \ourmethod, a unified end-to-end GraphRAG framework that tightly couples a GFM retriever with an LLM generator for joint retrieval and reasoning. As shown in the bottom of Figure~\ref{fig:intro}, \ourmethod fosters a reciprocal loop: the structural insights of GFM enhance LLM reasoning, while LLM reasoning refines graph retrieval. Given a query, the GFM retriever first reasons over the graph to score and retrieve top-$k$ relevant document nodes. Instead of directly feeding the retrieved documents into the LLM, we introduce a novel \emph{graph bridge} module to connect the GFM and LLM through a fully differentiable architecture. Specifically, the node embeddings of retrieved documents are extracted from the GFM and projected into ``graph prefixes'' that encode their structural roles and inter-document dependencies, and are then soft-weighted by the corresponding retrieval scores. These graph prefixes are concatenated with the raw document texts to form graph-augmented prompts for the LLM. In this way, the structural knowledge captured by the GFM is injected directly into the generation process, which enables the LLM to reason not only over the retrieved contents but also over their graph-structured dependencies (\textbf{Addressing Limitation 3}).

More importantly, the graph bridge remains fully differentiable, which allows gradients from the final answer generation loss to flow back through the graph prefixes and soft weights to refine both the LLM generator and GFM retriever. This directly aligns retrieval with downstream reasoning objectives and enables joint optimization of the two modules (\textbf{Addressing Limitation 2}). To further stabilize training, we introduce a \emph{semantic-guided distillation} module that provides auxiliary supervision for the GFM retriever using pseudo-labels derived from query-document semantic similarities. This module helps the retriever learn robust topological navigation from semantic priors without requiring expensive fine-grained human annotations (\textbf{Addressing Limitation 1}).

The core contributions of this work are summarized as follows:
\begin{itemize}
    \item We propose \ourmethod, a unified end-to-end GraphRAG framework that unifies a GFM retriever with an LLM generator, which enables joint optimization of retrieval and reasoning instead of the conventional decoupled retrieve-then-generate pipeline.
    \item We design a differentiable graph bridge that transforms retrieved document nodes into soft-weighted graph prefixes and graph-augmented prompts, which injects graph structure into the LLM while allowing answer supervision to back-propagate to the GFM retriever.
    \item We introduce a semantic-guided distillation module that stabilizes GFM training with pseudo-labels derived from query-document semantic similarity, which reduces reliance on costly fine-grained human annotations.
    \item Extensive experiments on three multi-hop QA datasets and three out-of-domain GraphRAG benchmarks show that \ourmethod consistently outperforms strong baselines and generalizes effectively across domains, which demonstrate the effectiveness of our unified framework.  
\end{itemize}

%% file: sections/02_related_work.tex
\section{Related Work}
In this section, we review some representative prior works for Graph Retrieval Augmented Generation (GraphRAG) and Graph Foundation Models (GFMs).
\subsection{Graph Retrieval Augmented Generation}
Graph retrieval-augmented generation has emerged as a promising paradigm that retrieves query-relevant graph elements (e.g., node, triples, paths, or subgraphs) from a pre-constructed graph database (e.g., knowledge graphs~\cite{luoreasoning,luo2025graph}). By explicitly modeling the inherent logical relationships among isolated pieces of knowledge, these methods significantly enhance the retrieval process beyond standard document retrieval. Recent approaches further integrate textual documents into knowledge graphs to facilitate more structured and semantically rich retrieval. For example, GraphRAG~\cite{edge2024local} employs community detection algorithms to summarize and organize documents, while LightRAG~\cite{guo2024lightrag} introduces a dual-level retrieval mechanism that combines structural graph retrieval with semantic vector retrieval to efficiently locate entities and their relations. Similarly, HippoRAG~\cite{jimenez2024hipporag} utilizes the Personalized PageRank (PPR) algorithm to identify salient graph entities and subsequently maps them to relevant documents. However, constructing these graphs is often computationally expensive and token-intensive, and the process can introduce noise, which limits the practical deployment of these methods in real-world scenarios. Concurrently, efforts to incorporate Graph Neural Networks (GNNs) into graph-enhanced RAG~\cite{he2024g,mavromatis2025gnn} have yielded expressive results, largely due to the multi-hop graph reasoning capabilities of GNNs over incomplete graphs~\cite{yasunaga2021qa,ren2024survey,he2025subgraph}. Nevertheless, these methods still struggle to generalize across diverse domains due to the absence of a unified graph foundation model.

\subsection{Graph Foundation Models}
% Graph Foundation Models and GFM-LLM
Graph Foundation Models (GFMs) leverage large-scale pre-training across massive datasets to create unified models with strong generalizability across various downstream tasks~\cite{liu2025graph,mao2024position,li2025advancing}. A primary challenge lies in defining graph tokens capable of capturing invariance across diverse and heterogeneous graph data~\cite{jiang2021pre}. For example, ULTRA~\cite{galkintowards} designs four fundamental relational interactions in knowledge graphs (KGs) to create a lightweight GFM with 0.2M parameters for link prediction. OpenGraph~\cite{xia2024opengraph} develops a graph tokenizer to convert graphs into a unified node token representation, which enables transformer-like GFMs for tasks such as link prediction and node classification. GFT~\cite{wang2024gft} introduces a transferable tree vocabulary to conduct a GFM that demonstrates effectiveness across diverse domains in graph learning. Despite these successful efforts, most methods primarily focus on conventional graph-related tasks and Transformer-like GFMs~\cite{tang2024higpt,tang2024graphgpt} often struggle to scale to large graphs. 

To combine the power of GFM and LLM, GraphLLM~\cite{chai2025graphllm} integrates graph learning models with LLM to equip LLM with the ability to reason on graph data. GraphGPT~\cite{tang2024graphgpt} also incorporates a text-graph grounding component to establish connections between text and graph structures for LLMs. GraphTranslator~\cite{zhang2024graphtranslator} bridges GFM and LLM with a translator, aiming to transfer the node representations into tokens, based on which LLM can perform node classification and link prediction.
Beyond conventional graph learning tasks, GFM-RAG~\cite{luo2025gfm} adopts a GFM-based retriever for complex question answering. However, this method relies heavily on the ground-truth query-document relevance labels for training the retriever and employs a disjoint two-stage pipeline. This inherently limits end-to-end optimization and often leads to sub-optimal reasoning performance. Therefore, designing a joint framework that seamlessly integrates a GFM-based retriever with a generator remains an open and critical question. %\zc{May add KKD/TKDE GFM methods and traditional end-to-end training methods (GFM-LLM).}

%% file: sections/04_approach.tex
\section{Approach}
\input{sections/tables/notations}

\begin{figure*}[t]
    \centering
    \includegraphics[width=1\textwidth]{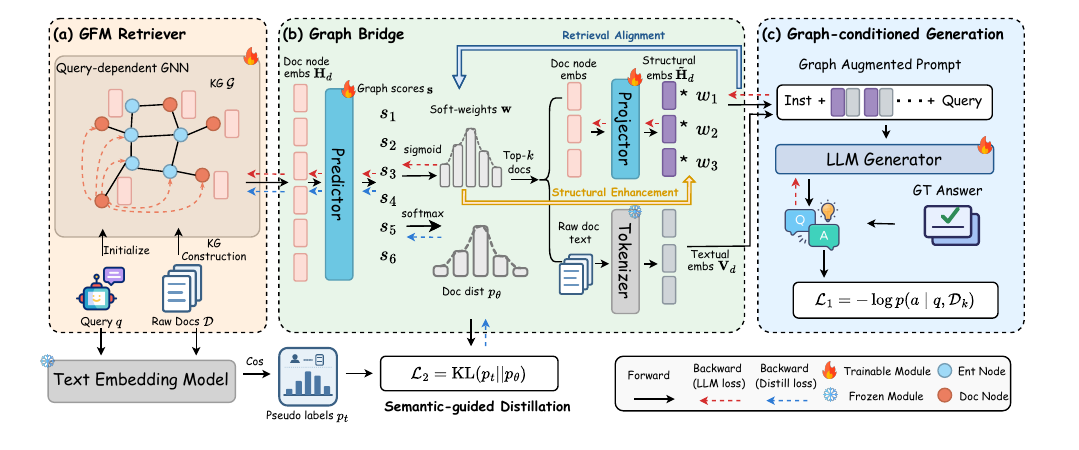}
    \caption{The overall framework of \ourmethod, which comprises of three core components: (a) \textit{a graph foundation model retriever}, which reasons over the structural knowledge for accurate document retrieval; (b) \textit{a graph bridge module}, which seamlessly connects the retriever and the generator into a unified framework; (c) \textit{a graph-conditioned LLM generator}, which incorporates the structural information to guide the final answer generation. In addition, we introduce a semantic-guided distillation module to provide auxiliary supervision for training GFMs.}
    \label{fig:framework}
\end{figure*}

The overall framework of \ourmethod is illustrated in Figure~\ref{fig:framework}, which consists of three core components: (a) a \emph{graph foundation model retriever} to effectively reason over the structured knowledge for document retrieval, (b) a \emph{graph bridge module} to connect the retriever and generator into a unified framework, and (c) a \emph{graph-conditioned LLM generator} that effectively incorporates graph information into the LLM generation process and enables end-to-end training. In the following sections, we will detail the formulation and design of each module.

\input{sections/03_preliminaries}

\subsection{Graph Foundation Model Retriever}\label{sec:gfm-retriever}
Graph foundation models (GFMs) \cite{mao2024position,liu2025graph} have demonstrated strong capabilities in capturing complex topological and semantic relationships within graph-structured data, which makes them ideal candidates for graph-based retrieval tasks. In \ourmethod, we leverage a GFM as the backbone of our retriever to effectively reason over the knowledge graph and identify query-relevant document nodes.

Specifically, we adopt a query-dependent Graph Neural Network (GNN) architecture~\cite{galkintowards,luo2025gfm} as the core of our GFM retriever. This design allows the model to explicitly incorporate both relational semantics and rich semantic features in node text into the message-passing process, which enables query-aware graph reasoning. 

\noindent\textbf{Text Encoding.} Given a graph $\mathcal{G}$, we first utilize a pre-trained text embedding model (e.g., Qwen3-Embedding~\cite{zhang2025qwen3}) to respectively encode the textual content associated with each node $v\in \mathcal{V}$ and each relation $r\in \mathcal{R}$ into initial $d$-dimensional node embeddings $\mathbf{h}_v \in \mathbb{R}^d$ and relation embeddings $\mathbf{h}_r \in \mathbb{R}^d$, which can be formulated as:
\begin{align}
    \mathbf{h}_v &= \operatorname{TextEmbedding}\left(\text{Text}(v)\right), v \in \mathcal{V}, \\
    \mathbf{h}_r &= \operatorname{TextEmbedding}\left(\text{Text}(r)\right), r \in \mathcal{R},
\end{align}
where $\text{Text}(\cdot)$ denotes the textual content associated with the nodes and relations. These semantic embeddings preserve the original textual knowledge, which provides a robust initialization for the subsequent GNN reasoning. Similarly, for a given user query $q$, we employ the same text embedding model to extract its dense representation $\mathbf{h}_q\in\mathbb{R}^d$, formulated as:
\begin{equation}
    \mathbf{h}_q = \operatorname{TextEmbedding}\left(q\right).
\end{equation}

By sharing the same encoding strategy, the query and graph elements are projected into the same latent semantic space. This approach not only enables generalization across different graphs and queries but also facilitates effective interaction between the query and the graph during GNN reasoning.

\noindent\textbf{Query-dependent GNN Reasoning.} After obtaining the initial embeddings, we apply an $L$-layer query-dependent GNN to jointly reason over text semantics and graph topology. The GNN updates the node representations iteratively by aggregating information from their neighbors, which allows the model to capture intricate dependencies among nodes and produce query-aware relevance scores for document retrieval. Formally, the GNN updates are defined as follows:
\begin{align}
    \mathbf{h}_v^{(0)} &= \operatorname{Init}\left(\mathbf{h}_v, \mathbf{1}_{v \in \mathcal{V}_q}  \mathbf{h}_q\right), \quad v \in \mathcal{V}, \\
    \mathbf{m}_v^{(l)} &= \scalebox{0.90}{$\displaystyle \operatorname{Agg}\left(\left\{\operatorname{Msg}\left(\mathbf{h}_v^{(l-1)}, \mathbf{h}_r^{(l)} , \mathbf{h}_{v^{\prime}}^{(l-1)}\right) \mid\left(v, r, v^{\prime}\right) \in \mathcal{E}\right\}\right)$}, \\
    \mathbf{h}_v^{(l)} &= \operatorname{Update}\left(\mathbf{h}_v^{(l-1)}, \mathbf{m}_v^{(l)}\right), \quad l \in[1, L], \label{eq:emb}
\end{align}
where $\mathbf{h}_v^{(l)}$ denotes the embedding of node $v$ at the $l$-th GNN layer. The \texttt{Init} operator initializes the node embeddings by a single-layer MLP, which fuses the original node embedding $\mathbf{h}_v$ and the query embedding $\mathbf{h}_q$ if the node $v$ belongs to the set of query-related nodes $\mathcal{V}_q$ (i.e., $\mathbf{1}_{v \in \mathcal{V}_q}$). \soc{Query-related node set $\mathcal{V}_q$ is constructed by semantic retrieval. Specifically, we encode the query and all graph nodes by an embedding model (e.g., Qwen3-Embedding-0.6B), compute their cosine similarities, and select the top-$k$ most similar entity and document nodes as query-related nodes.}

During the message-passing phase at each layer $l$, the \texttt{Msg} function employs DistMult~\cite{yang2015embedding} to compute the incoming messages from neighbors based on target node embedding $\mathbf{h}_v^{(l-1)}$, the source node embedding $\mathbf{h}_{v^{\prime}}^{(l-1)}$ and the relation embedding $\mathbf{h}_r^{(l)}$. These messages are subsequently pooled by the \texttt{Agg} function (e.g., sum aggregation). Finally, the \texttt{Update} function computes the target node embedding $\mathbf{h}_v^{(l)}$ by combining the aggregated messages $\mathbf{m}_v^{(l)}$ with its previous representation $\mathbf{h}_v^{(l-1)}$ with another MLP. Concurrently, relation embeddings $\mathbf{h}_r^{(l)}$ are updated with a layer-specific MLP, defined as $\mathbf{h}_r^{(l)}=\operatorname{MLP}(\mathbf{h}_r^{(l-1)})$.

\noindent\textbf{Document Retrieval.} By encoding the textual content of graph nodes and relations into dense embeddings and applying $L$-layer of GNN, the GFM retriever can capture intricate dependencies among nodes and produce query-aware relevance scores for document retrieval. The final query-aware node representation $\mathbf{h}_v^{(L)}$ is then used to compute a relevance score $s_v$ for document retrieval as follows:
\begin{equation}
    s_v=\operatorname{Predictor}\left(\mathbf{h}_v^{(L)}, \mathbf{h}_v, \mathbf{h}_q\right), \label{eq:score}
\end{equation}
% \lh{$z_v \to s_v$?}
where $\mathbf{h}_v$ and $\mathbf{h}_q$ denote the original text embeddings of node $v$ and query $q$, respectively, and the predictor is implemented as a two-layer MLP to produce a relevance score $s_v$ for each document node.

The top-$k$ document nodes with the highest relevance scores are selected as the retrieved documents for subsequent generation. Formally, the retrieved document set $\mathcal{D}_k$ is defined as:
\begin{equation}
    \mathcal{D}_k = \operatorname{Top-}k(s_d), d \in \mathcal{D}. 
\end{equation}
% \lh{K or k? Should be consistent across the paper.}
However, this hard selection operation is inherently non-differentiable, which prevents end-to-end optimization between the retriever and the generator and prevents the abundant structural information encoded by the GFM from being effectively utilized by the LLM generator. 

\subsection{Graph Bridge Module}\label{sec:graph_bridge}
To construct a differentiable bridge between the GFM retriever and the LLM generator, we introduce a novel \emph{graph bridge} module that transforms the retrieved document nodes into soft-weighted graph prefixes and seamlessly integrates them into the LLM generation process. The graph bridge serves as a continuous interface that allows the structural information from the GFM to be directly injected into the LLM while enabling gradients from the final answer loss to back-propagate to the GFM retriever for joint optimization.

We first relax the discrete retrieval scores $s_d$ into continuous soft-weights $w_d$ by passing them through a \texttt{Sigmoid} function, which preserves the independent relevance of each document without forcing competitive normalization. This can be formulated as:
\begin{equation}
    w_d = \frac{1}{1 + \exp(-s_d)}, d \in \mathcal{D}_k. \label{eq:weight}
\end{equation} These differentiable soft-weights explicitly quantify the importance of each retrieved document and allow the retriever to learn which documents maximize the generator's generation performance.

To inject structural information, we reuse the node representations of retrieved documents $\{\textbf{h}_d^{(L)}\mid d\in\mathcal{D}_k\}$ obtained from the final GNN layer in GFM retriever, which is defined in Eq.~\eqref{eq:emb}. These representations encapsulate the rich topological context of the document $d$ within the graph, which will be transformed into the \emph{graph prefixes} to guide the generation process of LLM.

A two-layer MLP (i.e., Projector) is first adopted as an alignment projector to map the structural representation into the semantic space of the LLM. Crucially, to allow the gradient to flow back to the retriever, we scale the projected representations by the differentiable soft-weighted $w_d$ derived via Eq.~\eqref{eq:weight}, which is formulated as follows:
\begin{equation}
    \tilde{\mathbf{h}}_d = w_d \cdot \operatorname{Projector}\left(\mathbf{h}_d^{(L)}\right), d \in \mathcal{D}_k.
\end{equation}
This soft-weighted structural representation $\tilde{\mathbf{h}}_d \in \mathbb{R}^{D_{llm}}$ with $D_{llm}$ dimensions serves as a special graph prefix, which explicitly encodes the structural information and relevance of the retrieved document to guide the LLM generation.

\subsection{Graph-conditioned LLM Generator}
\label{sec:generation}

By integrating the graph prefixes into the LLM generation process, we enable the LLM to effectively leverage the graph structure during reasoning, which significantly enhances its performance on complex tasks that require reasoning over multiple knowledge.

\noindent\textbf{Graph-augmented Prompt Construction:} For each retrieved document $d \in \mathcal{D}_k$, we first extract its textual content and tokenize it into a sequence of tokens. These tokens are then mapped to their dense embedding sequence $\mathbf{V}_d\in\mathbb{R}^{|d| \times D_{llm}}$ using the LLM's input embedding layer, where $|d|$ represents the sequence length of the document and $D_{llm}$ denotes the hidden dimension of the LLM. This process captures the raw semantic information of the retrieved documents, which serves as the primary context for generation.

Then, we concatenate the graph prefix $\tilde{\mathbf{h}}_d$ with the document embedding sequence $\mathbf{V}_d$ to form a graph-injected representation $\hat{\mathbf{V}}_d$ for each retrieved document, which is formulated as follows:
\begin{equation}
    \hat{\mathbf{V}}_d = \left[ \tilde{\mathbf{h}}_d \parallel \mathbf{V}_d \right], d \in \mathcal{D}_k, \hat{\mathbf{V}}_d \in \mathbb{R}^{(|d|+1) \times D_{llm}},
\end{equation}
where ``$\parallel$'' denotes sequence concatenation.

Finally, the formulated representations $\hat{\mathbf{V}}_d$ of all $k$ retrieved documents are concatenated with the initial user query sequence to form the graph-augmented prompt for the LLM generator. \soc{The detailed prompt is illustrated in Figure~\ref{fig:reasoning_prompt} in Appendix~\ref{sec:trainingprompt}}, where the \texttt{<graph\_prefix>} denotes the injected graph prefix $\tilde{\mathbf{h}}_d$.

\noindent\textbf{Graph-conditioned Generation:} Conditioned on this graph-augmented prompt, the LLM generator auto-regressively decodes the target answer $a = \{a_1, a_2, \dots, a_{|a|}\}$. The probability of generating the entire answer sequence is factorized as:
\begin{equation}
    p_\phi(a\mid q, \mathcal{D}_k) = \prod_{t=1}^{|a|} p_{\phi}(a_t \mid a_{<t}, q, \hat{\mathbf{V}}_{\mathcal{D}_k}),
\end{equation}
where $\hat{\mathbf{V}}_{\mathcal{D}_k}=\{\hat{\mathbf{V}}_d\}_{d\in\mathcal{D}_k}$.

% The architectural design of applying soft-weight strictly to the structural prefix ensures that the original language modeling capabilities of LLM remain undisturbed, while still establishing a fully differentiable bridge. This gating mechanism enables the gradients from the LLM loss to back-propagate to the GFM retriever's parameters $\theta$ through the structural token $\tilde{\mathbf{h}}_d$.

\noindent\textbf{End-to-end Joint Optimization:} During the training phase, the primary objective for the LLM generator is to minimize the standard negative log-likelihood (NLL) loss, denoted as $\mathcal{L}_1$:
\begin{equation}
    \mathcal{L}_1 = - \sum_{t=1}^{|a|} \log p_{\phi}(a_t \mid a_{<t}, q, \hat{\mathbf{V}}_{\mathcal{D}_k}).
\end{equation}

The applying graph prefix ensures that the original language modeling capabilities of LLM remain undisturbed, while still establishing a fully differentiable bridge, which allows gradients from the LLM loss to back-propagate to the GFM retriever's parameters $\theta$ through the prefix $\tilde{\mathbf{h}}_d$. %\lh{$L_1$ and $L_2$ is not aligned with the loss in the figure}. 

Building upon the chain rule, the gradient flow from each retrieved document explicitly splits into dual pathways as follows:
\begin{equation}\label{eq:gradient}
\frac{\partial \mathcal{L}_1}{\partial \theta} = \underbrace{ \frac{\partial \mathcal{L}_1}{\partial \tilde{\mathbf{h}}_d} \frac{\partial \tilde{\mathbf{h}}_d}{\partial w_d} \frac{\partial w_d}{\partial \theta} }_{\text{Soft-weighting Pathway}} + \underbrace{ \frac{\partial \mathcal{L}_1}{\partial \tilde{\mathbf{h}}_d} \frac{\partial \tilde{\mathbf{h}}_d}{\partial \mathbf{h}_d^{(L)}} \frac{\partial \mathbf{h}_d^{(L)}}{\partial \theta} }_{\text{Structural Pathway}}, d \in \mathcal{D}_k,
\end{equation}
where the two terms on the right-hand side denote the back-propagated gradient through the continuous weight $w_d$ and the structured embedding $\mathbf{h}_d^{(L)}$, respectively. 

The soft-weighting mechanism offers two main advantages for joint optimization. During the forward pass, $w_d$ acts as a continuous gate that scales the structural embedding and directs the attention of LLM to highly relevant documents while filtering out noise. In the backward pass, this pathway enables effective credit assignment to the retriever by explicitly measuring how changes in retrieval scores $s_d$ affect the final generation loss $\mathcal{L}_1$. As a result, the retriever learns which documents best enhance generator performance, promoting synergistic optimization between retrieval and generation. %\lh{As we highlight the importance of soft-weighting, we need to conduct ablation studies for this.}.

\subsection{Semantic-guided Distillation}
In real-world applications, obtaining fine-grained, ground-truth annotations for query-document relevance is prohibitively expensive and often infeasible. To overcome this strict absence of direct supervision, we draw inspiration from knowledge distillation~\cite{hinton2015distilling}. Specifically, we propose to train GFM through weak supervision by using the abundant unlabeled nodes on the graph. The pre-trained text embedding models~\cite{devlin2019bert} exhibit robust semantic reasoning capabilities, which allow them to effectively capture the underlying relevance between the query and documents. Therefore, we utilize a frozen, pre-trained text encoder as a teacher to generate semantic pseudo-labels for all document nodes $\mathcal{D}$. The pseudo-label probability for a document $d$ is formulated by applying a $\texttt{Softmax}$ function over the inner products of their semantic embeddings, which can be formulated as
% \begin{equation}
%     p_t(d \mid q, \mathcal{D}) = \frac{\exp(\mathbf{h}_d^\top \mathbf{h}_q)}{\sum_{d' \in \mathcal{D}} \exp(\mathbf{h}_{d'}^\top \mathbf{h}_q)},
% \end{equation}
\soc{
\begin{equation}
    p_t(d \mid q, \mathcal{D}) = \frac{\exp(\mathbf{h}_d^\top \mathbf{h}_q / \tau)}{\sum_{d' \in \mathcal{D}} \exp(\mathbf{h}_{d'}^\top \mathbf{h}_q / \tau)},
\end{equation}}
where $\textbf{h}_q$ and $\textbf{h}_d$ denote the original text embeddings of query $q$ and the document $d$, respectively. \soc{The temperature parameter $\tau$ controls the smoothness of the probability distribution. A smaller temperature sharpens the distribution and assigns larger probabilities to more relevant documents, which helps provide high-confidence supervision signals during distillation.}

Similarly, the student GFM retriever generates its predicted retrieval distribution $p_\theta(d \mid q, \mathcal{G})$ by applying the identical operation $\texttt{Softmax}$ over its predicted relevance scores $s_v$, which is defined as Eq.~\eqref{eq:score}. Finally, we optimize the GFM retriever by minimizing the Kullback-Leibler (KL) divergence between its predicted retrieval distribution and the semantically constructed pseudo-label distribution, which can be formulated as:
% \begin{equation}
%     \mathcal{L}_2 = \operatorname{KL} \Big(p_t(d \mid q, \mathcal{D}) \parallel p_\theta(d \mid q, \mathcal{G}) \Big).
% \end{equation}
\begin{equation}
\mathcal{L}_2 = \mathbb{E}_{d \sim p_t(d \mid q, \mathcal{D})} \left[ \log \frac{p_t(d \mid q, \mathcal{D})}{p_\theta(d \mid q, \mathcal{G})} \right].
\end{equation}
% \lh{D has been used for dimension. Maybe we can just use $\text{KL}$}.

\subsection{Training Strategy and Inference}
To unify the semantic reasoning of the LLM generator and the structural retrieval of the GFM retriever, we train the entire \ourmethod framework fully from end-to-end. The overall optimization objective is formulated as a linear combination of the generation loss $\mathcal{L}_1$ and the distillation loss $\mathcal{L}_2$:

\begin{equation}
    \mathcal{L} = \mathcal{L}_1 + \lambda \mathcal{L}_2,
\end{equation}
where $\lambda$ is a tunable hyperparameter that balances the primary generation task and auxiliary retrieval mechanism.

Through this joint optimization paradigm, these two modules synergize effectively: (1) $\mathcal{L}_2$ provides robust prior guidance for the retriever via semantic pseudo-labels, resolving the scarcity of annotated labels. (2) Concurrently, $\mathcal{L}_1$ offers task-specific gradients through our differentiable soft-weighting bridge, which enables fine-tuning the retriever to fetch document nodes that directly maximize the final answering accuracy.

%% file: sections/tables/notations.tex
\begin{table}[t]
\centering
\caption{Glossary of commonly used notations.}
\label{tab:notations}
\resizebox{0.95\columnwidth}{!}{%
\begin{tabular}{@{}lp{10cm}@{}}
\toprule
\textbf{Notation} & \textbf{Description} \\ \midrule
\multicolumn{2}{@{}l}{\textit{Knowledge Graph \& Query}} \\ \midrule
$\mathcal{G}$ & The heterogeneous knowledge graph \\
$v \in \mathcal{V}$ & A generic node $v$ in the node set $\mathcal{V}$ \\
$r \in \mathcal{R}$ & A relation type $r$ in the relation set $\mathcal{R}$ \\
$(v,r,v^{\prime}) \in \mathcal{E}$ & An edge $(v,r,v^{\prime})$ in the edge set $\mathcal{E}$ \\
$t_v \in \mathcal{T}$ & The node type of $v$, belonging to the type set $\mathcal{T}$ \\  
$q, a$ & The user query and the final answer \\ 
$d \in \mathcal{D}$ & A document $d$ within the document subset $\mathcal{D} \subset\mathcal{V}$ \\\midrule
\multicolumn{2}{@{}l}{\textit{Representations}} \\ \midrule
$D_{gnn}$ & The hidden dimensions of the GNN embeddings \\
$D_{llm}$ & The hidden dimensions of the LLM embeddings \\
$L$ & The total number of GNN layers \\
$\mathbf{h}_q, \mathbf{h}_d$ & Initial text embeddings of the query $q$ and document $d$ \\
$\mathbf{h}_v^{(l)}$ & The embedding of node $v$ at the $l$-th GNN layer \\
$\mathbf{m}_v^{(l)}$ & The aggregated message received by node $v$ at the $l$-th layer \\
$\mathbf{h}_d^{(L)}$ & The final graph embedding of document node \\
$\tilde{\mathbf{h}}_d$ & The structural prefix token projected from the graph embedding \\
$\mathbf{V}_d$ & Dense token embedding sequence of document $d$ \\
$\hat{\mathbf{V}}_d$ & The soft-weighted, graph-augmented prompt representation for $d$ \\ \midrule
\multicolumn{2}{@{}l}{\textit{Retrieval \& Distillation}} \\ \midrule
$\mathcal{D}_k$ & The subset of the top-$k$ retrieved document nodes \\
$s_d$ & The predicted discrete relevance score for document $d$ \\
$w_d$ & The continuous soft weight for document $d$, $w_d \in (0, 1)$ \\
$p_t$ & The semantic pseudo-label distribution generated by the teacher \\
$p_\theta$ & The predicted retrieval distribution by the GFM student \\ \midrule
\multicolumn{2}{@{}l}{\textit{Optimization}} \\ \midrule
$\lambda$ & Hyperparameter balancing the generation and distillation losses \\
$\theta, \phi$ & Trainable parameters of the GFM retriever and LLM generator \\
$\mathcal{L}_1$ & The negative log-likelihood (NLL) generation loss of the LLM \\
$\mathcal{L}_2$ & The semantic-guided KL divergence distillation loss \\
$\mathcal{L}$ & The overall optimization objective \\ \bottomrule
\end{tabular}%
}
\end{table}

%% file: sections/03_preliminaries.tex
% \section{Preliminaries}
% In this section, we introduce the key concepts in our paper and formally define our problem. The main notions are summarized in Table~\ref{tab:notations}.

% \subsection{GraphRAG Overview}
\noindent\textbf{GraphRAG Overview.} To contextualize our proposed framework, we begin by formally defining the general GraphRAG process. Given a set of documents $\mathcal{D}=\{d_1,d_2,\cdots,d_n\}$, following common GraphRAG approaches~\cite{edge2024local,luo2025gfm}, we employ LLMs to extract structural knowledge and construct a knowledge graph $\mathcal{G}=\left(\mathcal{V},\mathcal{E},\mathcal{R},\mathcal{T}\right)$. Here, $\mathcal{V}$ and $\mathcal{R}$ denote the sets of nodes and relations, respectively, and $\mathcal{E}=\{\left(v,r,v^{\prime}\right)\ \subseteq \mathcal{V}\times\mathcal{R}\times\mathcal{V}\}$ represents the set of edges that model the relationship among knowledge. Additionally, $\mathcal{T}$ specifies the node types (e.g., entities, documents). 

Given the constructed graph $\mathcal{G}$ and a user query $q$, standard GraphRAG paradigms first retrieve query-relevant information and then leverage an LLM to reason over the retrieved context and generate the final answer $a$. Formally, a graph retriever $R_{\theta}$ identifies the top-$k$ query-relevant document nodes $\mathcal{D}_k \subset \mathcal{D}$ according to the predicted probability $p_\theta(\mathcal{D}_k\mid q, \mathcal{G},\mathcal{D})$, and a generator $G_\phi$ then produces the final answer $a$ conditioned on both the query and the retrieved documents, which can be formulated as:
\begin{align}
\mathcal{G}&=\operatorname{GraphConstructor}\left(\mathcal{D}\right), \\
    p_{\Theta}(a\mid q, \mathcal{G},\mathcal{D})&=\sum_{\mathcal{D}_k \subseteq \mathcal{D}}p_\phi(a\mid q, \mathcal{D}_k)p_\theta(\mathcal{D}_k\mid q, \mathcal{G},\mathcal{D}),
\end{align}
where $\Theta = \{\theta, \phi\}$ denote the combined parameters of the retriever and generator.

%% file: sections/05_experiments.tex
\section{Experiments}\label{sec:exp}
In this section, we conduct extensive experiments to answer the following research questions: \textbf{RQ1:} Can \ourmethod achieve the state-of-the-art performance in complex reasoning? \textbf{RQ2:} Can \ourmethod effectively generalize across diverse domains and data? \textbf{RQ3:} How do the key components of \ourmethod contribute to overall performance? \textbf{RQ4:} How sensitive is \ourmethod to different hyper-parameter settings? \textbf{RQ5:} How does the choice of backbone model impact the effectiveness of \ourmethod? \textbf{RQ6:} How efficient is \ourmethod in terms of inference?

\subsection{Experimental Setup}
\noindent\textbf{Datasets.}
 To comprehensively evaluate \ourmethod, we conduct experiments on three widely adopted multi-hop QA datasets, including HotpotQA~\cite{yang2018hotpotqa}, MuSiQue~\cite{trivedi2022musique}, and 2WikiMultiHopQA (2Wiki)~\cite{ho2020constructing}. For the joint training phase, we utilize the training data curated by GFM-RAG~\cite{luo2025gfm}, where each dataset split consists of 1000 questions along with their corresponding documents. During the evaluation phase, we strictly follow the testing settings established by HippoRAG~\cite{jimenez2024hipporag}, GFM-RAG~\cite{luo2025gfm} for a fair comparison. We first assess the in-domain effectiveness of our proposed method \ourmethod on test datasets of the aforementioned three multi-hop QA benchmarks. Furthermore, to assess the zero-shot generalization capabilities of our framework across diverse domains, we employ three out-of-domain GraphRAG benchmarks: G-bench (Novel)~\cite{xiang2025use}, G-bench (Medical)~\cite{xiang2025use}, and G-bench (CS)~\cite{xiao2025graphrag}. These benchmarks rigorously test the complex reasoning skills of the model over medical, novel, and computer science (CS) knowledge without any domain-specific fine-tuning. The detailed statistics of the evaluation datasets are summarized in Table~\ref{tab:test_datasets}.

\noindent\textbf{Baseline Methods.}
To rigorously evaluate the superiority of \ourmethod, we compare it against a diverse set of strong baseline methods, which are broadly categorized into three groups. 
(1) \textit{Pure LLM Methods}: This category directly employs the pure LLM for answer generation without relying on external retrieval, including zero-shot (ZS) GPT-4o-mini~\cite{gpt4o} and Qwen3-8B~\cite{yang2025qwen3_tech} as well as a fine-tuned (FT) Qwen3-8B on the training data. (2) \textit{Traditional RAG Methods}: This category represents traditional sparse and dense retrieval models that rely solely on lexical or semantic matching without structural awareness, including BM25~\cite{robertson1994some}, ColBERTv2~\cite{santhanam2022colbertv2}, and Qwen3-Emb (e.g., Qwen3-Embedding-0.6B)~\cite{zhang2025qwen3}. 
(3) \textit{Graph-enhanced Methods}: This group consists of recent advancements that incorporate structural knowledge or graph reasoning into the retrieval and generation pipeline, including RAPTOR~\cite{sarthiraptor}, GraphRAG (MS)~\cite{edge2024local}, LightRAG~\cite{guo2024lightrag}, HippoRAG~\cite{jimenez2024hipporag}, SubgraphRAG~\cite{lisimple}, G-retriever~\cite{he2024g}, and GFM-RAG~\cite{luo2025gfm}. \soc{Detailed descriptions of these baseline methods are provided in Appendix~\ref{app}.} Crucially, since our realistic problem setting assumes no access to ground-truth labels for query-document relevance, we meticulously re-run the experiments for SubgraphRAG, G-retriever, and GFM-RAG under identical condition to ensure a fair comparison. Meanwhile, all baseline methods employ gpt-4o-mini as reasoning models, with the exception of Qwen3-8B (ZS) and Qwen3-8B (FT). Further, due to computational constraints, the experiments for \textbf{RQ4}, \textbf{RQ5}, and \textbf{RQ6} were performed on a sampled subset of the original training datasets.

\noindent\textbf{Metrics.}
We assess the performance of \ourmethod across two primary dimensions: final QA reasoning performance and intermediate retrieval performance. For QA reasoning performance, we report Exact Match (EM) and F1 scores, aligning with established practices in prior works~\cite{jimenez2024hipporag, luo2025gfm}. Conversely, for the G-bench datasets, we report standard Accuracy (ACC) to maintain strict consistency with their original benchmarks~\cite{xiang2025use, xiao2025graphrag}. Regarding retrieval performance, we adopt document recall@2 (R@2) and recall@5 (R@5) for multi-hop QA and evidence recall (Recall) for G-bench datasets as evaluation metrics.

\input{sections/tables/dataset}
\noindent\textbf{Implementation Details.}
We train \ourmethod using the large-scale corpus curated from GFM-RAG, which consists of \soc{60,000} query samples and 737,310 documents. For the architecture, the GFM retriever is implemented as a 6-layer query-dependent GNN with a hidden dimension of 1024. During message passing, we employ a DistMult function combined with sum aggregation, while the relation update mechanism is parameterized by a 2-layer MLP. The initial text representations are encoded using Qwen3-Embedding-0.6B. The total parameter of the retriever is 34M. The LLM generator is instantiated with the Qwen3-8B model.

Our proposed \ourmethod is trained with 8 NVIDIA H200 GPUs (141G) for 10 epochs with a per-device batch size 1. We optimize the framework using AdamW with differential learning rates to accommodate the varying convergence dynamics: 1e-4 for the randomly initialized GFM retriever and 2e-5 for the LLM generator. Finally, the loss-balancing hyperparameter $\lambda$ is set to 1. We set $k=5$ for multi-hop QA datasets and $k=10$ for G-bench datasets. \soc{We set the teacher temperature to $\tau_t=0.05$ to sharpen the teacher distribution and provide high-confidence relevance supervision, while the student temperature is set to $\tau_s=1.0$ during training.}

\subsection{Main Results \textbf{(RQ1)}}

\noindent\textbf{QA Results.} As shown in Table~\ref{tab:qa_multihop}, \ourmethod outperforms all baseline methods across three multi-hop question answering datasets, demonstrating its effectiveness in reasoning over graph-structured knowledge for complex reasoning. Pure LLM methods (e.g., GPT-4o-mini), even after fine-tuning (Qwen3-8B (FT)), perform poorly on these tasks due to deficiency in inherent knowledge and complex reasoning. Traditional RAG methods (e.g, BM25, Qwen3-Emb) mitigate knowledge gap by incorporating external knowledge, which generally outperform pure LLMs. However, they still struggle with multi-hop QA due to their inability to capture complex structural relationships. Graph-enhanced methods (e.g., HippoRAG, GFM-RAG) show improved performance by effectively leveraging graph structures. However, these methods either incur significant API cost (e.g, HippoRAG) or rely on high-quality query-document labels to train the retriever (e.g., GFM-RAG). 
Moreover, these two-stage methods lead to misalignment between the retriever and generator, resulting in suboptimal QA performance as the retrieved information may not be perfectly optimized for the downstream reasoning.

\ourmethod overcomes these limitations with a unified end-to-end framework that tightly integrates the GFM retriever and LLM generator, aligning their objectives to directly optimize for the final QA performance. By modeling complex relationships in the retrieved knowledge graph and ensuring that the retrieved information is highly relevant for reasoning, \ourmethod significantly outperforms all baselines and achieves the state-of-the-art across all evaluated datasets.

\input{sections/tables/qa_results}
\input{sections/tables/re_results}
\noindent\textbf{Retrieval Results.} As shown in Table~\ref{tab:retrieval_multihop}, \ourmethod achieves the best retrieval results on the MuSiQue and HotpotQA datasets, while maintaining highly competitive performance on the 2Wiki dataset. Although advanced embedding-based methods (e.g., Qwen3-Emb) yield strong results by effectively capturing semantic similarity, they consistently fall short of graph-enhanced approaches when handling complex, multi-hop queries. Notably, despite being distilled from the Qwen3-Emb model, \ourmethod significantly outperforms its semantic base model across all evaluated datasets. This substantial performance gap highlights a dual advantage in our approach. First, by utilizing the GFM to model topological relationships, \ourmethod successfully overcomes the inherent limitations of the semantic-only retrieval, proving that structural awareness is essential for complex reasoning. Second, our framework addresses the limitations of existing graph-based retrievers (e.g., GFM-RAG). Although GFM-RAG also adopts a GFM, \ourmethod still surpasses it by aligning the answer objectives directly into the retrieval phase. This demonstrates the superior efficacy of our end-to-end training paradigm over disjointed two-stage pipelines, ensuring that the retrieved information are perfectly optimized for the downstream reasoning task without relying on any human labeled ground-truth.
% \zc{GFM is better than text embedding model, which further demonstrate the effectiveness of the proposed method.}
\input{sections/tables/gbench}
\subsection{Generalization across Diverse Domains \textbf{(RQ2)}}
To evaluate the generalizability of \ourmethod, we conduct experiments on G-bench datasets in diverse domains, including novel, medical and computer science (CS). As shown in Table~\ref{tab:gbench}, \ourmethod consistently achieves top-tier performance in most metrics, which demonstrate strong robustness and cross-domain adaptability. Such superior generalization stems from the effective fusion of Graph Foundation Models and Large Language Models, which jointly leverage transferable graph representation ability and powerful reasoning and generalization capabilities. This synergy allows \ourmethod to perform reliably across distinct domains, which highlights its strong domain-agnostic nature.

\subsection{Ablation Study \textbf{(RQ3)}}
\begin{figure}[t]
    \centering
    \includegraphics[width=.9\columnwidth]{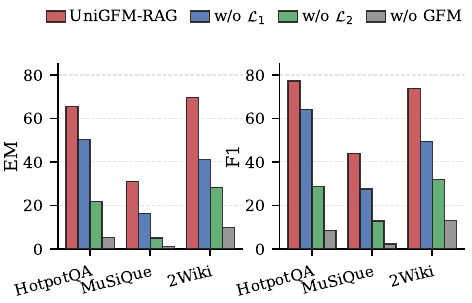}
    % \vspace{-.6cm}
    \caption{Ablation studies of \ourmethod.}
    \label{fig:ablation}
    % \vspace{-.6cm}
\end{figure}

In this section, we conduct ablation studies to validate the effectiveness of each core component in \ourmethod: (1) the LLM generator objective ($L_1$), (2) the semantic guided distillation loss ($L_2$), and (3) graph foundation model (GFM) retriever to the overall performance. The detailed results are summarized in Figure~\ref{fig:ablation}.

From the table, the experimental results clearly demonstrate the necessity of each component in \ourmethod. First, removing the generator objective (\textit{w/o} $\mathcal{L}_1$) yields a noticeable performance drop across all datasets, which confirms its essential role in ensuring high-quality answer generation. Secondly, excluding the semantic-guided distillation loss (\textit{w/o} $\mathcal{L}_2$) leads to severe degradation. This highlights that, in the absence of ground-truth document retrieval labels, our pseudo-label supervision is vital for effectively training GFM retriever. Crucially, we observe a catastrophic performance drop when removing the GFM retriever entirely (\textit{w/o} GFM). This underscores the absolute necessity of reasoning capability over the graph-structured knowledge for resolving complex reasoning.

% % \input{sections/tables/ablation}

\subsection{Parameter Analysis \textbf{(RQ4)}}
\begin{figure}[t]
    \centering
    \includegraphics[width=.85\columnwidth]{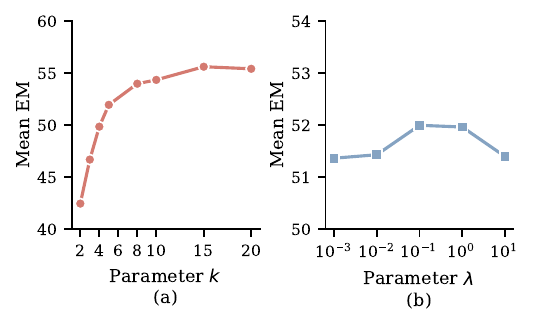}
    % \vspace{-.6cm}
    \caption{Parameter analysis on hyperparameter on $k$ and $\lambda$.}
    \label{fig:para}
    % \vspace{-.6cm}
\end{figure}

To investigate the robustness of \ourmethod, we analyze the impact of critical hyper-parameters, the number of the retrieved document $k \in \{2, 5, 8, 10, 15, 20\}$ and the distillation loss weight $\lambda \in \{0.001, 0.01,0.1,1.0,10\}$. As illustrated in Figure~\ref{fig:para}(a), while overall performance improves as $k$ increases, \ourmethod already achieves highly competitive results at $k=5$. Beyond this point, the marginal performance gains diminish rapidly, whereas the computational overhead grows significantly. Consequently, to strike an optimal trade-off between model effectiveness and computational efficiency, we adopt $k=5$ as the default setting. Additionally, Figure~\ref{fig:para}(b) demonstrates that the framework remains largely stable regarding $\lambda$. \ourmethod exhibits minor performance fluctuations under varying $\lambda$ values. Based on these observations, we set $\lambda=1.0$ as the optimal default setting for our experiments.

\input{sections/tables/backbone}
\subsection{Different Backbone Models \textbf{(RQ5)}}
To verify the model-agnostic nature of \ourmethod, we evaluate it with multiple LLM backbones of different families and parameter scales. As shown in Table~\ref{tab:backbone}, replacing the default generator with Qwen3-4B, Llama-3.1-8B, or Llama-3.2-3B still yields consistently strong performance across all three benchmarks. In particular, all backbone variants substantially outperform GFM-RAG on HotpotQA, and also achieve clear gains on the more challenging MuSiQue benchmark, indicating that the effectiveness of \ourmethod is not tied to a specific generator. On 2Wiki, the performance remains highly competitive, with the lightweight Llama-3.2-3B variant even achieving the best EM score among all compared backbones. These results show that the gains of \ourmethod mainly come from the framework itself, namely the tight integration of the GFM retriever and the LLM generator, rather than from relying on one particular backbone model.

\subsection{Efficiency Analysis \textbf{(RQ6)}}
\begin{figure}[t]
    \centering
    \includegraphics[width=0.75\columnwidth]{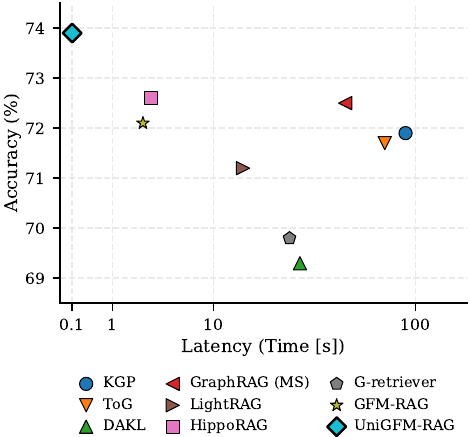}
    \vspace{-.1cm}
    \caption{Efficiency and performance comparison on G-bench (CS).}
    \label{fig:effi}
    \vspace{-.7cm}
\end{figure}

To evaluate the inference efficiency of \ourmethod, we conduct experiments on G-bench (CS) to compare both inference cost (seconds per sample) and reasoning accuracy against various baseline methods. As shown in Figure~\ref{fig:effi}, \ourmethod achieves the highest reasoning accuracy while also incurring the lowest inference time per sample. \soc{This advantage mainly comes from its end-to-end graph-aware reasoning design. Existing graph-based RAG methods usually involve additional online reasoning steps. For example, ToG requires multi-round graph exploration and iterative LLM prompting, while HippoRAG performs multi-hop graph retrieval with Personalized PageRank, which also introduces iterative graph propagation during inference. GraphRAG and LightRAG also rely on external LLMs to summarize, reason over, or generate responses from retrieved graph contexts, which increases the online inference cost. In contrast, \ourmethod performs retrieval and reasoning through a unified GFM, which avoids repeated external LLM invocations. Although graph encoding and message passing introduce additional computation, these operations are highly parallelizable on GPUs and are more efficient than multi-step LLM prompting during inference. Therefore, \ourmethod achieves a favorable balance between effectiveness and efficiency for reasoning over graph-structured knowledge.} % These results demonstrate that \ourmethod is Pareto-optimal in balancing effectiveness and efficiency for reasoning over graph-structured knowledge.

\input{figures/re_case}
\input{figures/qa_case}

\subsection{Retrieval Case Study}
To further illustrate the effectiveness of \ourmethod, we present a qualitative retrieval case study. As shown in Table~\ref{tab:re_case}, both \ourmethod and GFM-RAG successfully retrieve the initial relevant document for ``Rip Curl Pro Portugal'', which mentions John John Florence. However, GFM-RAG fails to retrieve the subsequent relevant information (e.g., date of birth) for ``John John Florence'', which directly derives the final answer. In contrast, \ourmethod successfully identifies the critical supporting document for ``John John Florence''. This demonstrates that, with the guidance of the LLM generator, our GFM retriever effectively navigates complex graph structures to locate critical multi-hop evidence for complex reasoning.

\subsection{QA Case Study}
Furthermore, we present a case study on answer generation to validate the critical impact of document ordering driven by our soft-weighting mechanism. As shown in Table~\ref{tab:qa_case}, both \ourmethod and GFM-RAG successfully retrieve all supporting documents within their top-5 results (Recall@5=1.0). However, \ourmethod places the relevant documents at the top of the ranked list, whereas GFM-RAG ranks them 3rd and 5th. Consequently, \ourmethod derives the correct answer, while GFM-RAG fails, likely as it suffers from the well-known ``lost in the middle'' phenomenon~\cite{baker2024lost}. Ultimately, this demonstrates that introducing the soft-weighting mechanism explicitly reveals the importance of each retrieved document, guiding the generator to focus on the most critical evidence.% \lh{Show the weights of each retrieved docs for our method in the table.}

%% file: sections/tables/dataset.tex
\begin{table}[t]
    \centering
    % \vspace{-.8cm}
    \caption{Statistics of the evaluation datasets.}
    \label{tab:test_datasets}
    \small
    \resizebox{.85\columnwidth}{!}{%
    \begin{tabular}{@{}lcc@{}}
        \toprule
        Dataset                                       & \# Query & \# Document \\ \midrule
        HotpotQA~\cite{yang2018hotpotqa}             & 1,000    & 9,221       \\
        MuSiQue~\cite{trivedi2022musique}            & 1,000    & 6,119       \\
        2Wiki~\cite{ho2020constructing}            & 1,000    & 11,656      \\
        G-bench (Novel)~\cite{xiang2025use}          & 2,010    & 461         \\
        G-bench (Medical)~\cite{xiang2025use}        & 2,062    & 2,406       \\
        G-bench (CS)~\cite{xiao2025graphrag}       & 1,018    & 24,534      \\ \bottomrule
    \end{tabular}%
    }
\end{table}

%% file: sections/tables/qa_results.tex
\begin{table}[t]
\centering
\caption{QA reasoning performance comparison on multi-hop datasets.}
\label{tab:qa_multihop}
\small
\resizebox{.75\columnwidth}{!}{%
\begin{tabular}{@{} l *{6}{c} @{}}
\toprule
\multirow{2}{*}{Method} & \multicolumn{2}{c}{HotpotQA} & \multicolumn{2}{c}{MuSiQue} & \multicolumn{2}{c}{2Wiki} \\
\cmidrule(lr){2-3} \cmidrule(lr){4-5} \cmidrule(l){6-7} 
& EM & F1 & EM & F1 & EM & F1 \\ 
\midrule
\multicolumn{7}{c}{\cellcolor[HTML]{EFEFEF}\textbf{Pure LLM Methods}} \\
GPT-4o-mini (ZS)~\cite{gpt4o}  & 28.6 & 41.0 & 11.2 & 36.3 & 30.2 & 36.3 \\
Qwen3-8B (ZS)~\cite{yang2025qwen3_tech}  & 15.1 & 23.7 & 1.3 & 8.1 & 23.2 & 27.9 \\
Qwen3-8B (FT)~\cite{yang2025qwen3_tech}  & 19.1 & 27.0 & 2.7 & 10.7 & 25.8 & 29.0 \\

\multicolumn{7}{c}{\cellcolor[HTML]{EFEFEF}\textbf{Traditional RAG Methods}} \\
\addlinespace
BM25~\cite{robertson1994some}    & 52.0 & 63.4 & 20.3 & 28.8 & 47.9 & 51.2 \\
ColBERTv2~\cite{santhanam2022colbertv2} & 43.4 & 57.7 & 15.5 & 26.4 & 33.4 & 43.3 \\
Qwen3-Emb~\cite{zhang2025qwen3} & 46.9 & 60.4 & 24.0 & 35.8 & 49.5 & 57.1 \\ 
\midrule
\multicolumn{7}{c}{\cellcolor[HTML]{EFEFEF}\textbf{Graph-enhanced Methods}} \\
\addlinespace
RAPTOR~\cite{sarthiraptor}       & 50.6 & 64.7 & 27.7 & 39.2 & 39.7 & 48.4 \\
GraphRAG (MS)~\cite{edge2024local} & 51.4 & 67.6 & 27.0 & 42.0 & 34.7 & 61.0 \\
LightRAG~\cite{guo2024lightrag}  & 9.9  & 20.2 & 2.0  & 9.3  & 2.5  & 12.1 \\
HippoRAG~\cite{jimenez2024hipporag} & 46.3 & 60.0 & 24.0 & 35.9 & 59.4 & 67.3 \\
SubgraphRAG~\cite{lisimple}      & 44.5 & 57.0 & 25.1 & 35.7 & 62.7 & 69.0 \\
G-retriever~\cite{he2024g}       & 41.4 & 53.4 & 23.6 & 34.3 & 33.5 & 39.6 \\
GFM-RAG~\cite{luo2025gfm}        & 46.3 & 60.2 & 23.5 & 34.8 & 60.0 & 66.8 \\ 
\midrule
\ourmethod  & \textbf{65.6} &\textbf{77.2} & \textbf{31.1} & \textbf{44.0} & \textbf{69.7} & \textbf{73.9} \\ 
\bottomrule
\end{tabular}%
}
\end{table}

%% file: sections/tables/re_results.tex
\begin{table}[t]
\centering
\caption{Retrieval performance comparison on multi-hop QA datasets. Recall@$k$ (R@$k$) is used as the evaluation metric.}
\label{tab:retrieval_multihop}
\small
\resizebox{.75\columnwidth}{!}{%
\begin{tabular}{@{}lcccccc@{}}
\toprule
\multicolumn{1}{l|}{}                            & \multicolumn{2}{c|}{HotpotQA}                      & \multicolumn{2}{c|}{MuSiQue}                       & \multicolumn{2}{c}{2Wiki}                          \\ \cmidrule(l){2-7} 
\multicolumn{1}{l|}{\multirow{-2}{*}{Method}} & R@2            & \multicolumn{1}{c|}{R@5}            & R@2            & \multicolumn{1}{c|}{R@5}            & R@2            & \multicolumn{1}{c}{R@5}            \\ \midrule
\multicolumn{7}{c}{\cellcolor[HTML]{EFEFEF}\textbf{Traditional RAG Methods}}                                                                                                                                                             \\
\multicolumn{1}{l|}{BM25 \cite{robertson1994some}}                         & 55.4          & \multicolumn{1}{c|}{72.2}          & 32.3          & \multicolumn{1}{c|}{41.2}          & 51.8          & \multicolumn{1}{c}{61.9}          \\
\multicolumn{1}{l|}{ColBERTv2 \cite{santhanam2022colbertv2}}                   & 64.7          & \multicolumn{1}{c|}{79.3}          & 37.9          & \multicolumn{1}{c|}{49.2}          & 59.2          & \multicolumn{1}{c}{68.2}          \\
\multicolumn{1}{l|}{Qwen3-Emb \cite{zhang2025qwen3}}              & 67.5          & \multicolumn{1}{c|}{80.0}          & 41.9          & \multicolumn{1}{c|}{55.7}          & 63.6         & \multicolumn{1}{c}{70.0}          \\ 
\midrule
\multicolumn{7}{c}{\cellcolor[HTML]{EFEFEF}\textbf{Graph-enhanced Methods}}                                                                                                                                                              \\
\multicolumn{1}{l|}{RAPTOR \cite{sarthiraptor}}                      & 58.1          & \multicolumn{1}{c|}{71.2}          & 35.7          & \multicolumn{1}{c|}{45.3}          & 46.3          & \multicolumn{1}{c}{53.8}          \\
\multicolumn{1}{l|}{GraphRAG (MS) \cite{edge2024local}}               & 58.3          & \multicolumn{1}{c|}{76.6}          & 35.4          & \multicolumn{1}{c|}{49.3}          & 61.6          & \multicolumn{1}{c}{77.3}          \\
\multicolumn{1}{l|}{LightRAG \cite{guo2024lightrag}}                    & 38.8          & \multicolumn{1}{c|}{54.7}          & 24.8          & \multicolumn{1}{c|}{34.7}          & 45.1          & \multicolumn{1}{c}{59.1}          \\
\multicolumn{1}{l|}{HippoRAG \cite{jimenez2024hipporag}}                    & 60.1          & \multicolumn{1}{c|}{78.5}          & 41.2          & \multicolumn{1}{c|}{53.2}          & 68.4          & \multicolumn{1}{c}{\textbf{87.0}} \\
\multicolumn{1}{l|}{SubgraphRAG \cite{lisimple}}                 & 57.6          & \multicolumn{1}{c|}{71.7}          & 37.7          & \multicolumn{1}{c|}{47.6}          & \textbf{70.2}          & \multicolumn{1}{c}{85.3}          \\
\multicolumn{1}{l|}{G-retriever \cite{he2024g}}                 & 51.8          & \multicolumn{1}{c|}{63.8}          & 35.4          & \multicolumn{1}{c|}{43.5}          & 61.0          & \multicolumn{1}{c}{66.6}          \\
\multicolumn{1}{l|}{GFM-RAG \cite{luo2025gfm}}                       & 56.1          & \multicolumn{1}{c|}{76.4}          & 37.0          & \multicolumn{1}{c|}{51.2}          & 63.5          & \multicolumn{1}{c}{82.4}          \\\midrule
\multicolumn{1}{l|}{\ourmethod}                  & \textbf{68.7} & \multicolumn{1}{c|}{\textbf{92.4}} & \textbf{45.5} & \multicolumn{1}{c|}{\textbf{67.3}} & 65.7 & \multicolumn{1}{c}{81.7} \\ \bottomrule
\end{tabular}%
}
\end{table}

%% file: sections/tables/gbench.tex
\begin{table}[t]
\centering
\caption{Performance comparison on the G-bench dataset\soc{~\cite{xiang2025use,xiao2025graphrag}}. QA reasoning is measured by Accuracy (ACC), and retrieval is measured by Recall.}
\label{tab:gbench}
\small
\resizebox{.85\columnwidth}{!}{%
\begin{tabular}{@{} l ccc cc @{}}
\toprule
\multirow{2}{*}{Method} & \multicolumn{3}{c}{G-bench (ACC)} & \multicolumn{2}{c}{G-bench (Recall)} \\
\cmidrule(lr){2-4} \cmidrule(l){5-6}
& Novel & Medical & CS & Novel & Medical \\ 
\midrule
\multicolumn{6}{c}{\cellcolor[HTML]{EFEFEF}\textbf{Pure LLM Methods}} \\
GPT-4o-mini (ZS)~\cite{gpt4o}  & 51.4 & 67.1 & 70.7 & - & - \\
Qwen3-8B (ZS)~\cite{yang2025qwen3_tech}  & 31.2 & 38.5 & 63.5 & - & - \\
Qwen3-8B (FT)~\cite{yang2025qwen3_tech}  & 42.2 & 39.4 & 24.9 & - & - \\

\multicolumn{6}{c}{\cellcolor[HTML]{EFEFEF}\textbf{Traditional RAG Methods}} \\
\addlinespace
BM25~\cite{robertson1994some}    & 56.5 & 68.7 & 71.7 & 82.1 & 87.9 \\
ColBERTv2~\cite{santhanam2022colbertv2} & 56.2 & 71.8 & 71.9 & 82.4 & 89.5 \\
Qwen3-Emb~\cite{zhang2025qwen3} & 46.0 & 40.5 & 73.4 & 86.4 & \textbf{95.2} \\ 

\midrule
\multicolumn{6}{c}{\cellcolor[HTML]{EFEFEF}\textbf{Graph-enhanced Methods}} \\
\addlinespace
RAPTOR~\cite{sarthiraptor}       & 43.2 & 57.1 & 73.6 & 66.1 & 84.2 \\
GraphRAG (MS)~\cite{edge2024local} & 50.9 & 45.2 & 72.5 & 67.4 & 56.4 \\
LightRAG~\cite{guo2024lightrag}  & 45.1 & 63.9 & 71.2 & 79.6 & 82.6 \\
HippoRAG~\cite{jimenez2024hipporag} & 44.8 & 59.1 & 72.6 & 81.2 & 84.0 \\
% SubgraphRAG~\cite{lisimple}      & -    & -    & -    & -    & -    \\
% G-retriever~\cite{he2024g}       & -    & -    & 69.8 & -    & -    \\
GFM-RAG~\cite{luo2025gfm}        & 56.7 & 71.9 & 72.6 & 83.5 & 89.6 \\ 

\midrule
\ourmethod  & \textbf{60.0} & \textbf{72.9} & \textbf{73.9} & \textbf{87.8} & 92.6 \\ 
\bottomrule
\end{tabular}%
}
\end{table}

%% file: sections/tables/backbone.tex
\begin{table}[t]
\centering
\caption{Different backbone models in \ourmethod.}
\label{tab:backbone}
\resizebox{.85\columnwidth}{!}{%
\begin{tabular}{@{}c|cccccc@{}}
\toprule
\multirow{2}{*}{Variant} & \multicolumn{2}{c}{HotpotQA} & \multicolumn{2}{c}{MuSiQue} & \multicolumn{2}{c}{2Wiki} \\ \cmidrule(l){2-7} 
                          & EM           & F1          & EM          & F1          & EM         & F1         \\ \midrule
GFM-RAG                & 46.3 & 59.5 & 23.5 & 34.8 & 60.0 & 66.8 \\ \midrule             
\multicolumn{7}{l}{\textit{\ourmethod}} \\                       
Qwen3-8B                & \textbf{63.4} &\textbf{75.7} & \textbf{31.1} & \textbf{44.0} & 61.7 & \textbf{67.1} \\
Qwen3-4B             & 61.2          & 73.2        & 26.7         & 37.5         & 59.3        & 64.67        \\
Llama-3.1-8B         & 59.5          & 72.3         & 29.4         & 39.2         & 59.0        & 63.5       \\
Llama-3.2-3B         & 58.0          & 70.8        & 28.3         & 39.9         & \textbf{62.4}        & 67.0        \\  \bottomrule
\end{tabular}%
}

\end{table}

%% file: figures/re_case.tex
% 请确保在导言区（preamble）引入了以下宏包：
% \usepackage{xcolor}
% \usepackage{pifont} % 用于打勾 \ding{51} 和打叉 \ding{55}

\begin{table*}[t]
    \centering
    \renewcommand{\arraystretch}{1.1} % 稍微增加行高，提升阅读舒适度
    \caption{Case study comparing the retrieval and reasoning performance of \ourmethod with GFM-RAG on the HotpotQA dataset.}
    \label{tab:re_case}
    \small 
    \resizebox{0.85\textwidth}{!}{%
    \begin{tabularx}{0.95\textwidth}{@{}l|X@{}}
        \toprule
        \textbf{Question}             & What American professional Hawaiian surfer born 18 October 1992 won the Rip Curl Pro Portugal? \\ 
        \textbf{Gold Answer}          & John John Florence \\ 
        \textbf{Supporting Docs.}     & [``MEO Rip Curl Pro Portugal'', ``John John Florence''] \\ 
        \midrule

        \multicolumn{2}{c}{\textbf{GFM-RAG (Baseline Method)}} \\ 
        \midrule
        \textbf{Retrieved Docs.}      & [``MEO Rip Curl Pro Portugal'', \textcolor{red!80!black}{``Reef Hawaiian Pro''}] \hfill \textbf{(Recall@2=0.5)} \\ 
        \textbf{Predicted Answer}     & document does not contain information about John John Florence \hfill \textcolor{red!80!black}{\ding{55} \textbf{Incorrect}} \\ 
        \midrule   
                
        \multicolumn{2}{c}{\textbf{\ourmethod (Ours)}} \\ 
        \midrule
        \textbf{Retrieved Docs.}      & [``MEO Rip Curl Pro Portugal'', \textcolor{teal}{\textbf{``John John Florence''}}] \hfill \textbf{(Recall@2=1.0)} \\ 
        \textbf{Predicted Answer}     & John John Florence \hfill \textcolor{teal}{\ding{51} \textbf{Correct}} \\ 
        \bottomrule
    \end{tabularx}}
\end{table*}
% \begin{table*}[t]
%     \centering
%     \caption{Case study comparing the retrieval and reasoning performance of \ourmethod and GFM-RAG on the HotpotQA dataset.}
%     \label{tab:re_case}
%     \small % 使用小一号字体使表格更紧凑
%     \begin{tabularx}{0.95\textwidth}{@{}l|X@{}}
%         \toprule
%         \textbf{Question}             & What American professional Hawaiian surfer born 18 October 1992 won the Rip Curl Pro Portugal? \\ \midrule
%         \textbf{Gold Answer}          & John John Florence \\ \midrule
%         \textbf{Supporting Docs.}     & [``MEO Rip Curl Pro Portugal'', ``John John Florence''] \\ \midrule

%         \multicolumn{2}{c}{\textbf{GFM-RAG (Baseline Method)}} \\ \midrule
%         \textbf{Retrieved Docs.}      & [``MEO Rip Curl Pro Portugal'', ``Reef Hawaiian Pro''] \textbf{(Recall@2=0.5)} \\ 
%         \textbf{Predicted Answer}     & document does not contain information about John John Florence \textbf{(Incorrect)} \\ \midrule   
                
%         \multicolumn{2}{c}{\textbf{\ourmethod (Ours)}} \\ \midrule
%         \textbf{Retrieved Docs.}      & [``MEO Rip Curl Pro Portugal'', ``John John Florence''] \textbf{(Recall@2=1)} \\ 
%         \textbf{Predicted Answer}     & John John Florence \textbf{(Correct)} \\ 
%         \bottomrule
%     \end{tabularx}
% \end{table*}

%% file: figures/qa_case.tex
\begin{table*}[t]
    \centering
    \renewcommand{\arraystretch}{1.1} % 增加行高，提升呼吸感
    \caption{Case study comparing the retrieval and reasoning performance of \ourmethod with GFM-RAG on the HotpotQA dataset.}
    \label{tab:qa_case}
    \small 
    \resizebox{0.85\textwidth}{!}{%
    \begin{tabularx}{0.95\textwidth}{@{}l|X@{}}
        \toprule
        \textbf{Question}             & Which season did Brian Billick coach the Baltimore Ravens though in 2001? \\ 
        \textbf{Gold Answer}          & 6th season \\ 
        \textbf{Supporting Docs.}     & [``2001 Baltimore Ravens season'', ``Brian Billick''] \\ 
        \midrule

        \multicolumn{2}{c}{\textbf{GFM-RAG (Baseline Method)}} \\ 
        \midrule
        \textbf{Retrieved Docs.}      & [``2015 Baltimore Ravens season'', ``2011 Baltimore Ravens season'', \textcolor{red!80!black}{``Brian Billick''}, ``2000 Baltimore Ravens season'', \textcolor{red!80!black}{``2001 Baltimore Ravens season''}] \hfill \textbf{(Recall@5=1.0)} \\ 
        \textbf{Predicted Answer}     & 2001 Baltimore Ravens season \hfill \textcolor{red!80!black}{\ding{55} \textbf{Incorrect}} \\ 
        \midrule
        \multicolumn{2}{c}{\textbf{\ourmethod (Ours)}} \\ 
        \midrule
        \textbf{Retrieved Docs.} & [\textcolor{teal}{\textbf{``2001 Baltimore Ravens season''}} (0.42), \textcolor{teal}{\textbf{``Brian Billick''}} (0.38), ``2000 Baltimore Ravens season'' (0.34), ``2015 Baltimore Ravens season'' (0.19), ``2001 St. Louis Rams season'' (0.03)] \hfill \textbf{(Recall@5=1.0)} \\
        \textbf{Predicted Answer}     & 6th season \hfill \textcolor{teal}{\ding{51} \textbf{Correct}} \\ 
        \bottomrule
    \end{tabularx}}
\end{table*}
% \begin{table*}[t]
%     \centering
%     \caption{Case study comparing the retrieval and reasoning performance of \ourmethod with GFM-RAG on the HotpotQA dataset.}
%     \label{tab:qa_case}
%     \small % 使用小一号字体使表格更紧凑
%     \begin{tabularx}{0.95\textwidth}{@{}l|X@{}}
%         \toprule
%         \textbf{Question}             & Which season did Brain Billick coach the Baltimore Ravens though in 2001? \\ \midrule
%         \textbf{Gold Answer}          & 6th season \\ \midrule
%         \textbf{Supporting Docs.}     & [``Gasland'', ``Pumping Iron''] \\ \midrule

%         \multicolumn{2}{c}{\textbf{GFM-RAG (Baseline)}} \\ \midrule
%         \textbf{Retrieved Docs.}      & [``2015 Baltimore Ravens season'', ``2011 Baltimore Ravens season'', ``Brian Billick'', ``2000 Baltimore Ravens season'', ``2001 Baltimore Ravens season'']  (Recall@5=1)\\ 
%         \textbf{Predicted Answer}     & no \textbf{(Incorrect)} \\ \midrule
                
%         \multicolumn{2}{c}{\textbf{\ourmethod (Ours)}} \\ \midrule
%         \textbf{Retrieved Docs.}      & [``2001 Baltimore Ravens season'' - 0.28, ``Brian Billick'' - 0.23, ``2000 Baltimore Ravens season'', ``2015 Baltimore Ravens season'', ``2001 St. Louis Rams season''] (Recall@5=1) \\ 
%         \textbf{Predicted Answer}     & 6th season \textbf{(Correct)} \\ 
%         \bottomrule
%     \end{tabularx}
% \end{table*}

%% file: sections/06_conclusions.tex
\section{Conclusion}
In this paper, we present \ourmethod, a unified, end-to-end GraphRAG framework that seamlessly couples a graph foundation model (GFM) retriever with a large language model (LLM) generator. By leveraging GFM, we effectively reason over structural knowledge to conduct accurate document retrieval. Through the proposed graph bridge module, we connect the retriever and generator via a fully differentiable architecture. This integration fosters a reciprocal learning loop: the structural insights from GFM enhance the reasoning capabilities of LLM, while the gradient feedback from the LLM refines the graph retrieval process. Furthermore, by employing soft-weighting mechanism, we enable the LLM to dynamically focus on the most critical documents within the graph-augmented prompt. Extensive experiments on six complex reasoning benchmarks demonstrate that \ourmethod consistently outperforms existing baseline methods, which establishes new state-of-the-art performance. Additional analyses confirm its exceptional inference efficiency and strong generalizability. Ultimately, \ourmethod offers a promising paradigm for unifying GFMs and LLMs for knowledge-intensive applications. \soc{In future work, we plan to extend \ourmethod in two directions. First, we will explore multi-modal knowledge injection, which enables the topology-aware reasoning process to incorporate heterogeneous information beyond pure text. Second, we aim to adapt the unified end-to-end framework to specialized and high-stakes domains that require both structural evidence and reliable generation.}

%% file: sections/biography.tex
\section{Biography Section}
\begin{IEEEbiography}[{\includegraphics[width=1in,height=1.25in,clip,keepaspectratio]{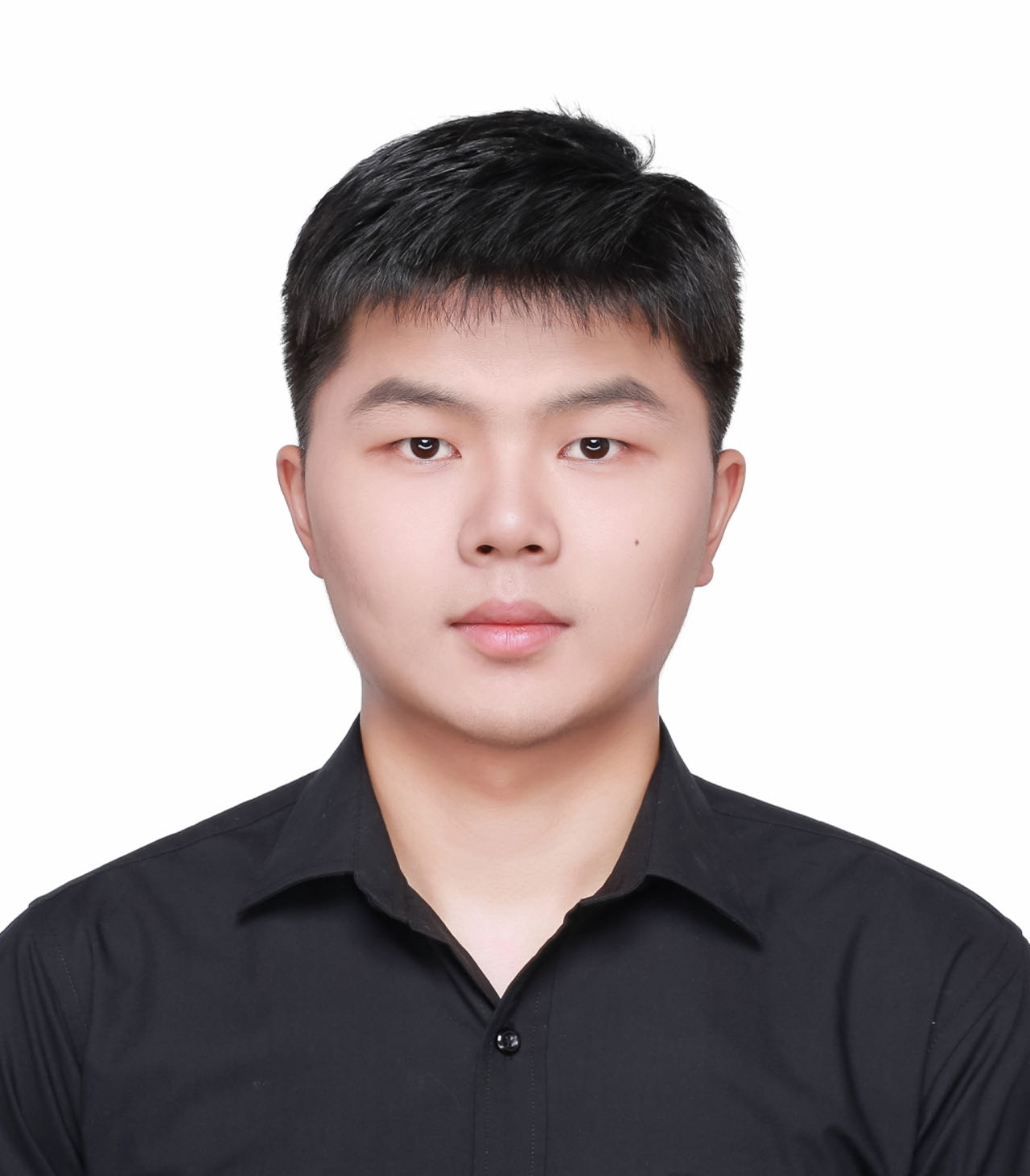}}]{Zicheng Zhao}
is currently pursuing the Ph.D. degree in computer science and technology. He is currently a visiting student with the School of Automation and Intelligent Sensing, Shanghai Jiao Tong University. His research interests include knowledge graphs, large language models, and graph neural networks.
\end{IEEEbiography}
\vspace{-10mm}
\begin{IEEEbiography}[{\includegraphics[width=1in,height=1.25in,clip,keepaspectratio]{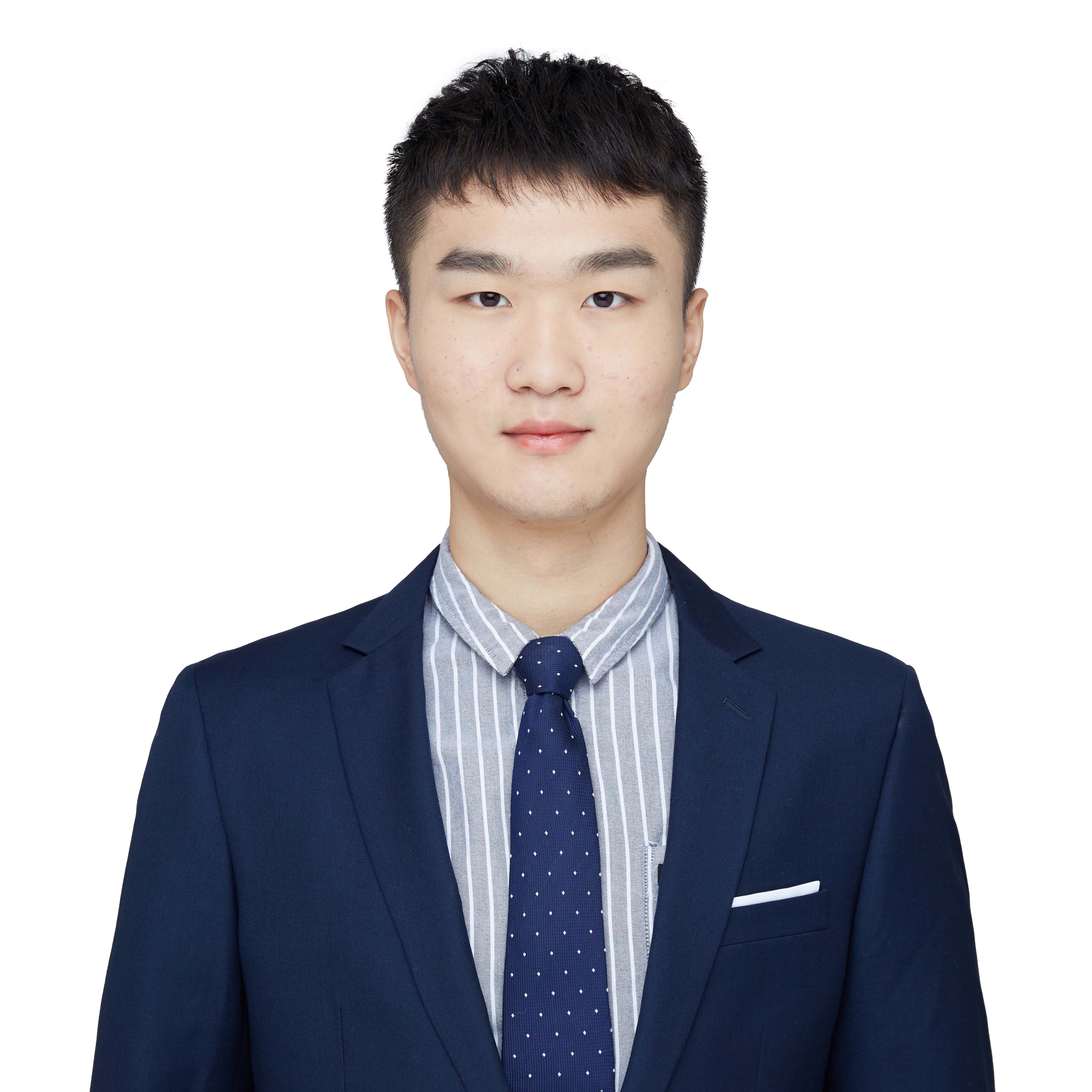}}]{Linhao Luo}
is a Postdoctoral Research Fellow with the Department of Data Science \& AI at Monash University, Australia. He obtained his PhD degree in computer science at Monash University and received his B.S. degree from Harbin Institute of Technology, China. He has published over 30 research papers in top-tier journals and conferences, including the IEEE Transactions on Knowledge and Data Engineering (TKDE), Knowledge and Information Systems, Knowledge-Based Systems, ICLR, ICML, NeurIPS, ACL, EMNLP, AAAI, IJCAI, WSDM, CIKM, SIGIR, and The Web Conference.
\end{IEEEbiography}
\vspace{-10mm}
\begin{IEEEbiography}[{\includegraphics[width=1in,height=1.25in,clip,keepaspectratio]{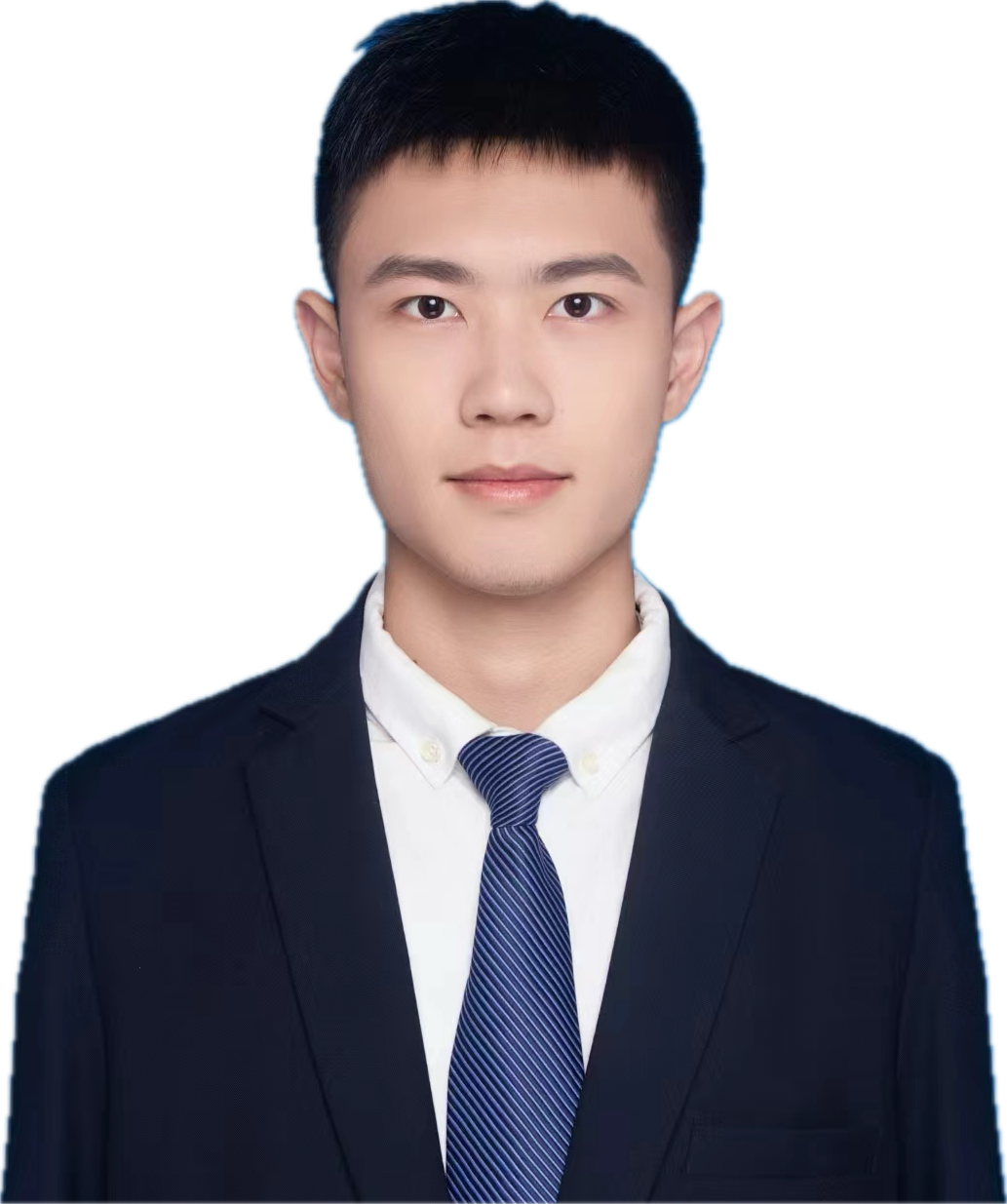}}]{Zongji Yu}
will receive the bachelor's degree from the School of Control Science and Engineering, Shandong University in Jinan, China. Since 2026, he has been pursuing a doctoral degree at Shanghai Jiao Tong University in Shanghai, China. His research interests include continual learning, reinforcement learning, and AI safety.
\end{IEEEbiography}
\vspace{-10mm}
\begin{IEEEbiography}[{\includegraphics[width=1in,height=1.25in,clip,keepaspectratio]{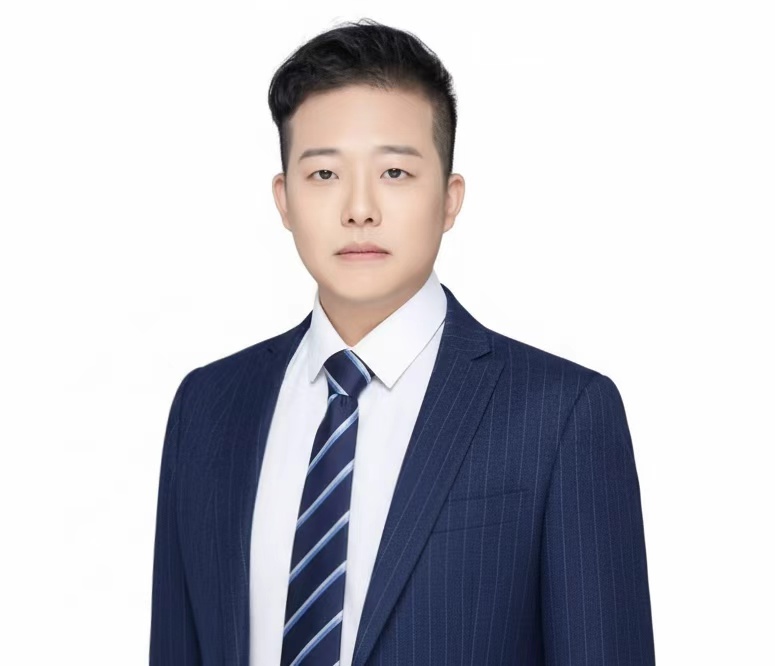}}]{Junnan Dong}
is now a senior researcher at Tencent Youtu Lab. He obtained his Ph.D. at the Hong Kong Polytechnic University. He focuses on Large Language Models (LLMs), Diffusion Language Models (DLMs), Retrieval Augmented Generation (RAG) and Agents for advanced domain-specific reasoning. He has published more than 20 papers on top-tier conferences, including ICLR, ICML, NeurIPS, ACL, WWW, KDD, SIGIR, AAAI and WSDM. He was also nominated as the 'Best Research Postgraduate Student' by both the Faculty and Department of Computing at PolyU. He has been continuously serving as a program committee member for ICLR, NeurIPS, ICML, ICLR, ACL, AAAI and WWW, etc.
\end{IEEEbiography}
\vspace{-10mm}
\begin{IEEEbiography}[{\includegraphics[width=1in,height=1.25in,clip,keepaspectratio]{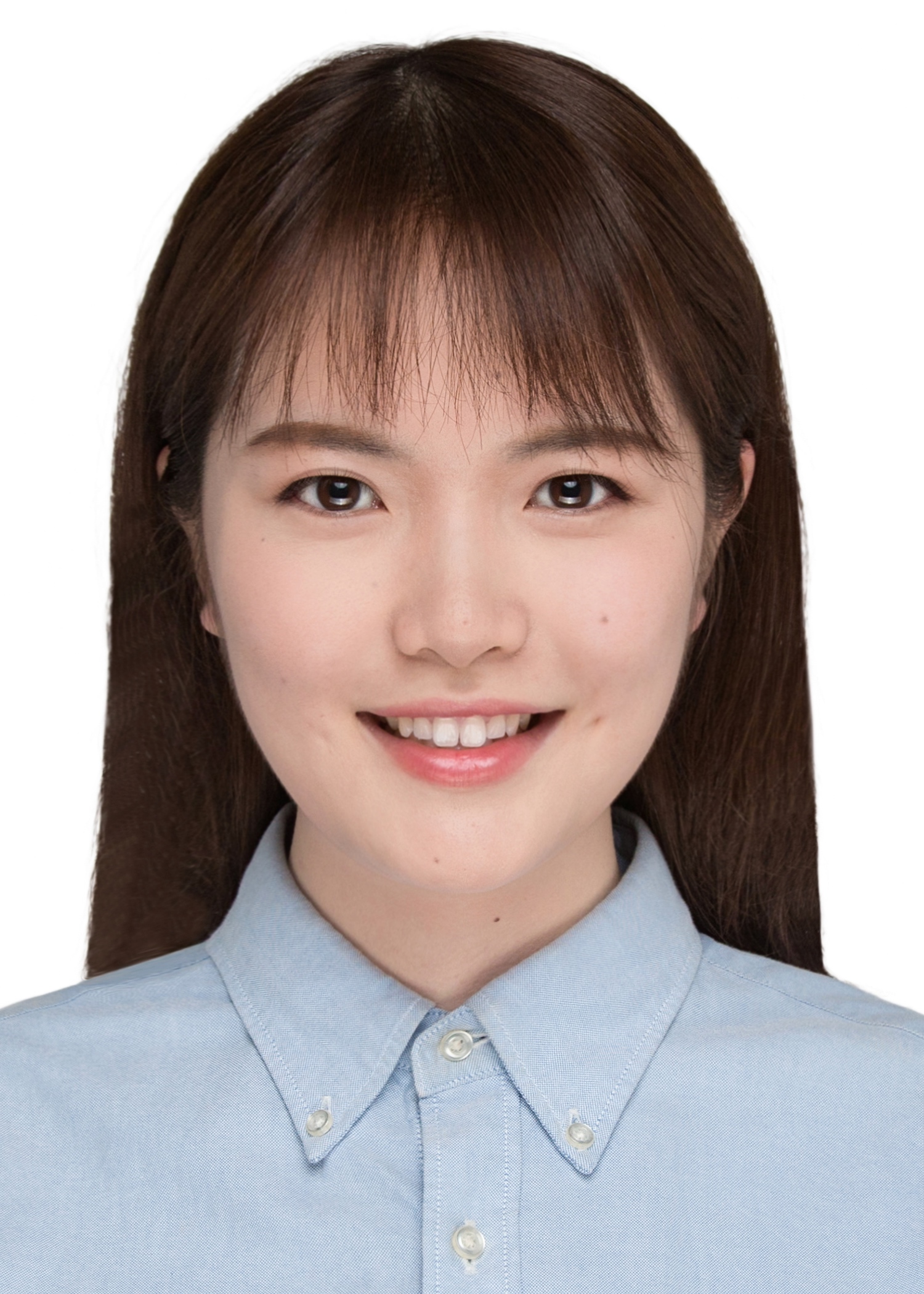}}]{Yanchao Tan} (Member, IEEE) is currently an Associate Professor in the College of Computer and Data Science, Fujian Key Laboratory of Network Computing and Intelligent Information Processing, Fuzhou University. She obtained her PhD degree in the College of Computer Science, Zhejiang University, China. Her research interests include data mining, healthcare, graph mining, and recommender system.
\end{IEEEbiography}
\vspace{-10mm}
\begin{IEEEbiography}[{\includegraphics[width=1in,height=1.25in,clip,keepaspectratio]{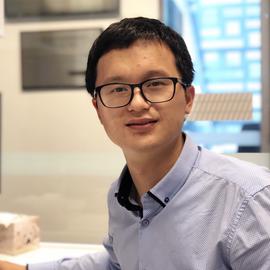}}]{Shirui Pan} (Senior Member, IEEE) received the PhD degree in computer science from the University of Technology Sydney (UTS), Ultimo, NSW, Australia. He is a professor with the School of Information and Communication Technology, Griffith University, Australia. His research interests include data mining and machine learning. His research received the 2024 CIS IEEE TNNLS Oustanding Paper Award and the 2020 IEEE ICDM Best Student Paper Award. He is recognized as one of the AI 2000 AAAI/IJCAI Most Influential Scholars in Australia. He was the awardee for the 2024 IEEE ICDM Tao Li Award. He is an ARC Future fellow and a fellow of Queensland Academy of Arts and Sciences (FQA).
\end{IEEEbiography}
\vspace{-10mm}
\begin{IEEEbiography}[{\includegraphics[width=1in,height=1.25in,clip,keepaspectratio]{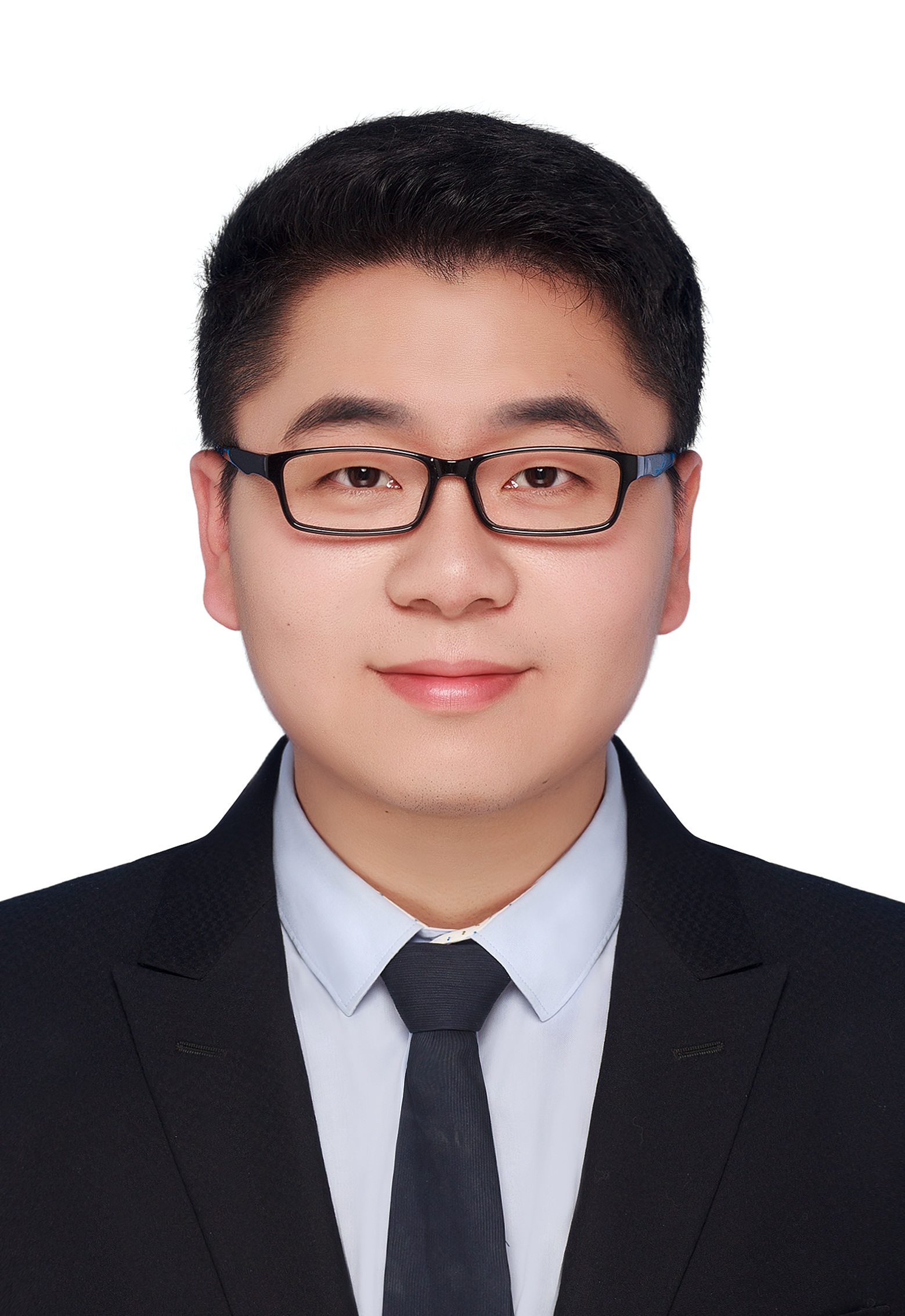}}]{Chen Gong} (Senior Member, IEEE)
received his dual doctoral degree from Shanghai Jiao Tong University (SJTU) and University of Technology Sydney (UTS), respectively. Currently, he is a full professor of Shanghai Jiao Tong University. His research interests mainly include machine learning, data mining, and learning-based vision problems. He has published more than 130 technical papers at prominent journals and conferences such as IEEE T-PAMI, JMLR, IJCV, IEEE T-NNLS, IEEE T-IP, ICML, NeurIPS, ICLR, CVPR, ICCV, ECCV, AAAI, IJCAI, ICDM, etc. He serves as the associate editor for IEEE T-CSVT, Neural Networks, and NePL, and also the Area Chair or Senior PC member of several top-tier conferences such as AAAI, IJCAI, ICML, ICLR, ECML-PKDD, AISTATS, ICDM, ACM MM, etc. He won the ICDM Best Student Paper Runner-Up Award, the second prize of Natural Science Award of the Chinese Institute of Electronics, ``Excellent Doctorial Dissertation Award'' of Chinese Association for Artificial Intelligence, ``Wu Wen-Jun AI Excellent Youth Scholar Award'', and the Scientific Fund for Distinguished Young Scholars of Jiangsu Province. He was also selected as the ``Global Top Chinese Young Scholars in AI'' released by Baidu, and “World's Top 2\% Scientists” released by Stanford University.
\end{IEEEbiography}

%% file: sections/07_appendix.tex
\section{Appendix}
\subsection{Statistics for Training Datasets}
The training data contains 60 KGs with over 14M triples constructed from 700k documents extracted from the training set. The statistics are shown in Table~\ref{tab:training_data}.
\input{sections/tables/training_statistic}

\subsection{Detailed Descriptions of Baseline Methods.}
\label{app}
To rigorously evaluate the effectiveness of \ourmethod, we compare it with a diverse set of strong baseline methods, which are broadly categorized into three groups. (1) \textit{Pure LLM Methods}: zero-shot (ZS) GPT-4o-mini~\cite{gpt4o} and Qwen3-8B~\cite{yang2025qwen3_tech} as well as a fine-tuned (FT) Qwen3-8B on the training data. (2) \textit{Traditional RAG Methods}: BM25~\cite{robertson1994some}, ColBERTv2~\cite{santhanam2022colbertv2}, and Qwen3-Emb (e.g., Qwen3-Embedding-0.6B)~\cite{zhang2025qwen3}. (3) \textit{Graph-enhanced Methods}: RAPTOR~\cite{sarthiraptor}, GraphRAG (MS)~\cite{edge2024local}, LightRAG~\cite{guo2024lightrag}, HippoRAG~\cite{jimenez2024hipporag}, SubgraphRAG~\cite{lisimple}, G-retriever~\cite{he2024g}, and GFM-RAG~\cite{luo2025gfm}.

\noindent\textbf{Pure LLM Methods} directly employ the LLMs for answer generation without relying on external retrieval.

\begin{itemize}
\item \textbf{GPT-4o-mini}~\cite{gpt4o}: A strong proprietary LLM used in the zero-shot setting, where the model directly generates answers without external retrieval.
\item \textbf{Qwen3-8B}~\cite{yang2025qwen3_tech}: An open-source LLM used in both zero-shot and fine-tuned settings to evaluate the capability of pure LLM-based answer generation.
\end{itemize}

\noindent\textbf{Traditional RAG Methods} represent traditional sparse and dense retrieval models that rely solely on lexical or semantic matching without structural awareness.

\begin{itemize}
\item \textbf{BM25}~\cite{robertson1994some}: A traditional probabilistic retrieval model that ranks documents according to exact term matching and term-frequency statistics.
\item \textbf{ColBERTv2}~\cite{santhanam2022colbertv2}: A dense retrieval model that improves retrieval quality and efficiency through late interaction, denoised training, and residual compression.
\item \textbf{Qwen3-Emb}~\cite{zhang2025qwen3}: A dense embedding retriever based on Qwen3-Embedding-0.6B, which retrieves documents according to semantic similarity between the query and candidate documents.
\end{itemize}

\noindent\textbf{Graph-enhanced Methods} consist of recent approaches that incorporate structural knowledge or graph reasoning into the retrieval and generation pipeline.

\begin{itemize}
\item \textbf{RAPTOR}~\cite{sarthiraptor}: A hierarchical retrieval framework that recursively clusters, embeds, and summarizes text chunks at different levels to support multi-level retrieval.
\item \textbf{GraphRAG (MS)}~\cite{edge2024local}: A graph-based RAG framework that constructs a corpus-level graph, applies hierarchical community detection, and generates community summaries to support LLM-based reasoning.
\item \textbf{LightRAG}~\cite{guo2024lightrag}: A graph-enhanced RAG framework that incorporates graph structures into text indexing and performs dual-level retrieval to capture both fine-grained entity information and high-level relational knowledge.
\item \textbf{HippoRAG}~\cite{jimenez2024hipporag}: A training-free graph retriever that performs multi-hop retrieval over a document-based knowledge graph using Personalized PageRank.
\item \textbf{SubgraphRAG}~\cite{lisimple}: A retrieval framework that extracts dynamically sized subgraphs through lightweight scoring mechanisms, enabling efficient graph-based retrieval without task-specific fine-tuning.
\item \textbf{G-retriever}~\cite{he2024g}: A graph-based RAG method that formulates graph retrieval as a Prize-Collecting Steiner Tree optimization problem to extract relevant knowledge from large textual graphs beyond the context window of standard LLMs.
\item \textbf{GFM-RAG}~\cite{luo2025gfm}: A graph foundation model based RAG framework that leverages structural reasoning and supports zero-shot retrieval on unseen datasets.
\end{itemize}

\subsection{Detailed Prompt for Training}\label{sec:trainingprompt}
To ensure reproducibility and clarify how the generator utilizes the retrieved graph-aware context during training, we provide the detailed LLM reasoning prompt in Fig.~\ref{fig:reasoning_prompt}. The prompt contains the retrieved documents preceded by graph-aware prefixes, which encode the structural information produced by \ourmethod. The LLM is then required to answer the question based only on the provided context and output the final answer without additional explanations. This design ensures that the generation process is conditioned on both textual evidence and graph-derived structural signals.
\input{figures/prompt}

\subsection{Fine-grained Analysis by Reasoning Depth}
To further evaluate the reasoning capability of \ourmethod under different levels of complexity, we provide a fine-grained analysis by grouping questions according to their required reasoning depth, which include 2-hop, 3-hop, and 4-hop cases. We report both EM and F1 scores to examine whether \ourmethod can maintain its advantage when more complex multi-hop reasoning paths are required.

\begin{figure}[t]
    \centering
    \includegraphics[width=.9\columnwidth]{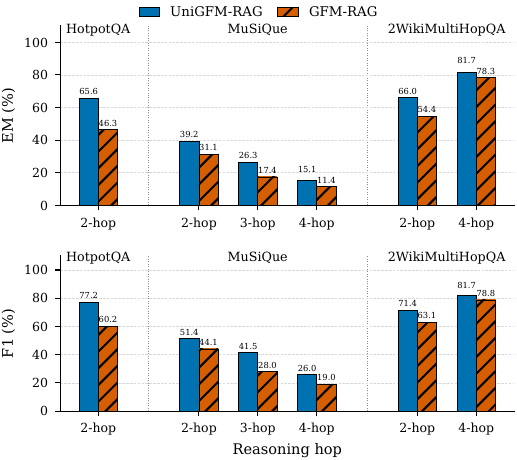}
    % \vspace{-.6cm}
    \caption{Fine-grained comparison of \ourmethod and GFM-RAG under different reasoning depths. EM and F1 scores are reported for 2-hop, 3-hop, and 4-hop questions across datasets.}
    \label{fig:complex}
    % \vspace{-.6cm}
\end{figure}

As shown in Figure~\ref{fig:complex}, \ourmethod consistently outperforms GFM-RAG across all datasets and reasoning depths. This demonstrates that the proposed unified framework is not only effective for relatively simple 2-hop questions, but also remains robust when the reasoning process involves longer and more complex evidence chains. More importantly, the advantage of \ourmethod is still clear in high-depth scenarios. For example, on the challenging MuSiQue dataset, \ourmethod substantially improves the F1 score over GFM-RAG on both 3-hop questions, from 28.0 to 41.5, and 4-hop questions, from 19.0 to 26.0. These results indicate that the end-to-end joint optimization in \ourmethod helps reduce error propagation across reasoning steps and better captures task-relevant topological reasoning paths.

\subsection{Complex Reasoning Case Study}
\input{figures/appendx_34hops}
To further examine the multi-hop reasoning capability of \ourmethod, we provide qualitative case studies on 3-hop and 4-hop questions from the challenging MuSiQue dataset. As shown in Tables~\ref{tab:3hopcase} and~\ref{tab:4hopcase}, \ourmethod exhibits a stronger ability to retrieve complete evidence chains than the decoupled baseline GFM-RAG, especially when the required reasoning path becomes longer and more complex.

For the 3-hop case in Table~\ref{tab:3hopcase}, GFM-RAG suffers from evidence loss and retrieves only one of the three supporting documents, which results in a Recall@5 of $1/3$ and an incorrect prediction. In contrast, \ourmethod successfully retrieves all three supporting documents and obtains a Recall@5 of $1.0$, which provides sufficient evidence for predicting the correct answer, ``Oriole Records''.

For the 4-hop case in Table~\ref{tab:4hopcase}, the reasoning challenge becomes more severe because the model needs to follow a longer evidence chain. GFM-RAG retrieves mostly irrelevant contexts and only recalls one of the four supporting documents, which leads to a Recall@5 of $1/4$ and an incorrect answer. By contrast, \ourmethod accurately traces the complete four-step reasoning path, retrieves all required supporting documents with a Recall@5 of $1.0$, and predicts the correct answer, ``Paraguay''. These qualitative results show that the unified end-to-end design of \ourmethod can better preserve evidence across extended topological reasoning paths, help reduce error propagation, and support more reliable answer generation in complex multi-hop scenarios.

\subsection{Token Cost for Graph Construction}
Token consumption during knowledge graph construction is an important factor for evaluating the deployment cost of graph-enhanced RAG methods. Since \ourmethod adopts the same knowledge graph construction pipeline as GFM-RAG, it does not introduce additional token consumption beyond this pipeline during the indexing stage. Table~\ref{tab:token} reports the token consumption required for index and knowledge graph construction per 1,000 documents.
\input{sections/tables/tokenconsume}

As shown in the table, \ourmethod requires approximately 4.8M tokens per 1,000 documents for graph construction, which is lower than LightRAG and GraphRAG. This lower construction cost makes \ourmethod more suitable for large-scale graph-enhanced RAG scenarios, especially when the corpus size grows and the upfront indexing cost becomes a major deployment consideration.

\subsection{Knowledge Graph Construction Prompt}
To improve the reproducibility of the knowledge graph construction process, we provide the OpenIE prompt used for extracting RDF-style triples in Fig.~\ref{fig:openie}. \ourmethod adopts the KG construction pipeline used in HippoRAG, where named entities are first extracted from each passage and then used to condition relation extraction. The prompt design follows Fig.~9 in the HippoRAG appendix~\cite{jimenez2024hipporag}.
\input{sections/tables/openie}

\subsection{Discussion}\label{sec:discuss}
This section further discusses two practical issues of \ourmethod that are important for real-world deployment and robust training. We first analyze the scalability of \ourmethod when applied to large-scale graph-structured knowledge sources, which focuses on parallel computation and subgraph-level reasoning. We then discuss the potential noise introduced by similarity-based weak supervision and explain how the proposed training framework can alleviate its negative impact through task-level feedback from the generation objective.

\noindent\textbf{Scalability to Large-scale Graphs.} We discuss the scalability of \ourmethod to large-scale graphs. First, the graph encoding and message-passing operations in \ourmethod are naturally parallelizable and can be partitioned across multiple GPUs. Distributed message-passing strategies, such as those adopted in G-reasoner, can also be incorporated to further scale the computation to larger graphs. Second, in practical RAG scenarios, it is often unnecessary to reason over the entire graph. \ourmethod can be combined with subgraph extraction, where a query-relevant local subgraph is first retrieved and then processed by the graph foundation model. This strategy reduces memory consumption and inference cost while preserving task-relevant structural information. Therefore, although \ourmethod introduces additional graph-oriented components, its parallelizable computation and compatibility with subgraph-level reasoning make it scalable to large graph-structured knowledge sources.

\noindent\textbf{Impact of Noisy Weak Supervision.} \ourmethod uses similarity-based signals as weak supervision because ground-truth supporting documents or reasoning paths are often unavailable in real-world multi-hop QA scenarios. Such weak supervision may inevitably contain noise, since semantically similar documents may not always be useful for answer generation. In our framework, the distillation loss $\mathcal{L}_2$ transfers the similarity-based relevance signal to the retriever and improves the recall of potentially relevant evidence. The ablation results in Figure~\ref{fig:ablation} show that removing $\mathcal{L}_2$ leads to consistent performance degradation across datasets, which confirms the necessity of this weak relevance supervision. Meanwhile, the end-to-end training framework provides task-level feedback from the generation objective to adjust document importance. As shown in Eq.~\eqref{eq:gradient}, gradients from the generation loss can be propagated to retrieved documents through both the soft-weighting and structural pathways, which helps  the model emphasize answer-supporting documents and reduce the influence of irrelevant ones. These results suggest that \ourmethod can benefit from similarity-based weak supervision while reducing the negative impact of noisy weak supervision through end-to-end optimization.

%% file: sections/tables/training_statistic.tex
\begin{table}[h]
    \centering
    \caption{Statistics of the query-doc pairs and KGs used for training.}
    \label{tab:training_data}
    \resizebox{\columnwidth}{!}{%
        \begin{tabular}{@{}ccccccc@{}}
            \toprule
            Dataset & \#Q-doc Pair & \#Document & \#KG & \#Entity & \#Relation & \#Triple  \\ \midrule
            HotpotQA & 20,000        & 204,822    & 20    & 1,930,362  & 967,218     & 6,393,342  \\
            MuSiQue  & 20,000        & 410,380    & 20    & 1,544,966  & 900,338     & 4,848,715  \\
            2Wiki    & 20,000        & 122,108    & 20    & 916,907    & 372,554     & 2,883,006  \\
            \midrule
            \textbf{Total} & \textbf{60,000} & \textbf{737,310} & \textbf{60} & \textbf{4,392,235} & \textbf{2,240,110} & \textbf{14,125,063} \\ \bottomrule
        \end{tabular}%
    }
\end{table}

%% file: figures/prompt.tex
\begin{figure}[t]
    \centering
    \begin{minipage}{0.85\columnwidth}
        \centering
        \begin{tcolorbox}[title=LLM Reasoning Prompt]
            \small
            \begin{lstlisting}[breaklines=true, breakindent=0pt, breakatwhitespace=false, basicstyle=\ttfamily]
You are a helpful assistant that answers questions based on the provided context. Please only respond with the answer and no additional text.

### Context:
<graph_prefix> document_text_1
<graph_prefix> document_text_2
...
<graph_prefix> document_text_n

### Question:
<Question>
            \end{lstlisting}
        \end{tcolorbox}
        \vspace{1mm}
    \end{minipage}
    \caption{The graph-augmented prompt for LLM Reasoning.}
    \label{fig:reasoning_prompt}
\end{figure}

%% file: figures/appendx_34hops.tex
\begin{table*}[t]
    \centering
    \renewcommand{\arraystretch}{1.3} % 稍微增加行高，提升阅读舒适度
    \caption{Case study comparing the retrieval and reasoning performance of \ourmethod with GFM-RAG on the MuSiQue dataset (\textbf{3-hop}).}
    \label{tab:3hopcase}
    \small 
    \begin{tabularx}{0.95\textwidth}{@{}l|X@{}}
        \toprule
        \textbf{Question}             & What UK label was bought by the broadcast company that, along with ABC and the original network of Undercovers, is one of the major broadcasters based in New York? \\ 
        \textbf{Gold Answer}          &  Oriole Records \\ 
        \textbf{Supporting Docs.}     & [``New York City\_5'', ``Undercovers (TV series)'', ``Sony Music\_4''] \\ 
        \midrule

        \multicolumn{2}{c}{\textbf{GFM-RAG (Baseline Method)}} \\ 
        \midrule
        \textbf{Retrieved Docs.}      & [\textcolor{red!80!black}{``Undercovers (TV series)''}, ``Neil Reid'', ``999 ABC Broken Hill'', ``Julie (TV series)'', ``The Night Shift (season 3)''] \hfill \textbf{(recall@5=1/3)} \\ 
        \textbf{Predicted Answer}     & ITV \hfill \textcolor{red!80!black}{\ding{55} \textbf{Incorrect}} \\ 
        \midrule   
                
        \multicolumn{2}{c}{\textbf{\ourmethod (Ours)}} \\ 
        \midrule
        \textbf{Retrieved Docs.}      & [\textcolor{teal}{\textbf{``Undercovers (TV series)''}}, \textcolor{teal}{\textbf{``New York City\_5''}}, \textcolor{teal}{\textbf{``Sony Music\_4''}}, ``Roxy Recordings'', ``New York Giants Radio Network''] \hfill \textbf{(recall@5=1.0)} \\ 
        \textbf{Predicted Answer}     & Oriole Records \hfill \textcolor{teal}{\ding{51} \textbf{Correct}} \\ 
        \bottomrule
    \end{tabularx}
\end{table*}

\begin{table*}[h]
    \centering
    \renewcommand{\arraystretch}{1.3} % 稍微增加行高，提升阅读舒适度
    \caption{Case study comparing the retrieval and reasoning performance of \ourmethod with GFM-RAG on the MuSiQue dataset (\textbf{4-hop}).}
    \label{tab:4hopcase}
    \small 
    \begin{tabularx}{0.95\textwidth}{@{}l|X@{}}
        \toprule
        \textbf{Question}             & A country's military branch, the equivalent of which in the US helped the Old West's local police, was unprepared for the invasion of the country that developed the ZH-29. The unprepared country was the only communist country to have an embassy where? \\ 
        \textbf{Gold Answer}          &  Paraguay \\ 
        \textbf{Supporting Docs.}     & [``Police'', ``ZH-29'', ``Josip Broz Tito\_1'', ``Josip Broz Tito\_3''] \\ 
        \midrule

        \multicolumn{2}{c}{\textbf{GFM-RAG (Baseline Method)}} \\ 
        \midrule
        \textbf{Retrieved Docs.}      & [``International Women's Day'', \textcolor{red!80!black}{``ZH-29''}, ``Czech Republic'', ``International Day of the Girl Child'', ``1973 oil crisis\_1''] \hfill \textbf{(recall@5=1/4)} \\ 
        \textbf{Predicted Answer}     & United States \hfill \textcolor{red!80!black}{\ding{55} \textbf{Incorrect}} \\ 
        \midrule   
                
        \multicolumn{2}{c}{\textbf{\ourmethod (Ours)}} \\ 
        \midrule
        \textbf{Retrieved Docs.}      & [\textcolor{teal}{\textbf{``ZH-29''}}, \textcolor{teal}{\textbf{``Josip Broz Tito\_1''}}, ``Warsaw Pact'', \textcolor{teal}{\textbf{``Josip Broz Tito\_3''}}, \textcolor{teal}{``Police''}] \hfill \textbf{(recall@5=1.0)} \\ 
        \textbf{Predicted Answer}     & Paraguay \hfill \textcolor{teal}{\ding{51} \textbf{Correct}} \\ 
        \bottomrule
    \end{tabularx}
\end{table*}

%% file: sections/tables/tokenconsume.tex
\begin{table}[h]
\centering
\caption{Token consumption for index and knowledge graph construction per 1,000 documents.}
\label{tab:token}
\begin{tabular}{lc}
\toprule
\textbf{Method} & \textbf{Token Consumption} \\
\midrule
\ourmethod & 4.8M \\
LightRAG & 5.5M \\
GraphRAG & 7.6M \\
\bottomrule
\end{tabular}
\end{table}

%% file: sections/tables/openie.tex
\begin{figure*}[t]
\centering
\begin{minipage}{0.96\textwidth}
\begin{tcolorbox}[title=Open Information Extraction Prompt, colback=white, colframe=black!60]
\scriptsize
\begin{lstlisting}[breaklines=true, breakindent=0pt, breakatwhitespace=false, basicstyle=\ttfamily\scriptsize]
Instruction:
Your task is to construct an RDF (Resource Description Framework) graph from the given passages and named entity lists.
Respond with a JSON list of triples, with each triple representing a relationship in the RDF graph.
Pay attention to the following requirements:
- Each triple should contain at least one, but preferably two, of the named entities in the list for each passage.
- Clearly resolve pronouns to their specific names to maintain clarity.

Convert the paragraph into a JSON dict, it has a named entity list and a triple list.

One-Shot Demonstration:
Paragraph:

Radio City
Radio City is India's first private FM radio station and was started on 3 July 2001.
It plays Hindi, English and regional songs.
Radio City recently forayed into New Media in May 2008 with the launch of a music portal - PlanetRadiocity.com that offers music related news, videos, songs, and other music-related features.

{"named_entities":
["Radio City", "India", "3 July 2001", "Hindi", "English", "May 2008", "PlanetRadiocity.com"]
}

Assistant:
{"triples": [
["Radio City", "located in", "India"],
["Radio City", "is", "private FM radio station"],
["Radio City", "started on", "3 July 2001"],
["Radio City", "plays songs in", "Hindi"],
["Radio City", "plays songs in", "English"],
["Radio City", "forayed into", "New Media"],
["Radio City", "launched", "PlanetRadiocity.com"],
["PlanetRadiocity.com", "launched in", "May 2008"],
["PlanetRadiocity.com", "is", "music portal"],
["PlanetRadiocity.com", "offers", "news"],
["PlanetRadiocity.com", "offers", "videos"],
["PlanetRadiocity.com", "offers", "songs"]
]}

Input:

Convert the paragraph into a JSON dict, it has a named entity list and a triple list.
Paragraph:
```
PASSAGE TO INDEX
```
{"named_entities": [NER LIST]}
\end{lstlisting}
\end{tcolorbox}
\end{minipage}
\caption{Open information extraction prompt for knowledge graph construction.}
\label{fig:openie}
\end{figure*}